%% file: main.tex
\pdfoutput=1
\documentclass[10pt]{article} % For LaTeX2e
\usepackage[accepted]{tmlr}
\input{math_commands.tex}

\usepackage{hyperref}
\usepackage{url}
\usepackage{algorithm}
\usepackage{algorithmic}
\usepackage{newfloat}
\usepackage{listings}
\usepackage{wrapfig}
\let\AND\relax

\usepackage{makecell}
\usepackage{subfigure}
\usepackage{enumitem}
\usepackage{amssymb}
\usepackage{nicefrac}
\usepackage{amsthm}
\usepackage{amsmath}
\usepackage{booktabs}
\usepackage{multirow}
\usepackage{graphicx}
\usepackage{capt-of}
\usepackage{placeins}
\usepackage{multicol}
\usepackage{xcolor}

\newtheorem{theorem}{Theorem}
\newtheorem{proposition}{Proposition}
\newtheorem{lemma}{Lemma}
\newtheorem{definition}{Definition}

\usepackage{pifont}  % Required for \ding command

\title{Graph Unitary Message Passing}

\author{\name Haiquan Qiu$^1$ \email qhq22@mails.tsinghua.edu.cn \\
\name Quanming Yao$^{1,2,3}$ \email qyaoaa@tsinghua.edu.cn \\
      \addr $^1$Department of Electrical Engineering, Tsinghua University \\
      \AND
      \addr $^2$Beijing National Research Center for Information Science and Technology \\
      \AND
      \addr $^3$State Key laboratory of Space Network and Communications
}

\def\month{06}  % Insert correct month for camera-ready version
\def\year{2026} % Insert correct year for camera-ready version
\def\openreview{\url{https://openreview.net/forum?id=dvNMDkSBIA}} % Insert correct link to OpenReview for camera-ready version

\begin{document}

\maketitle

\begin{abstract}

Unitarity is a useful principle for stabilizing deep neural networks, but in graph neural networks (GNNs) instability is induced not only by learnable parameters but also by the graph propagation operator. Motivated by this distinction, we propose Graph Unitary Message Passing (GUMP), a message-passing framework that uses a unitary propagation operator on a transformed graph to avoid graph-induced exponential decay under repeated propagation. GUMP combines (i) a graph transformation that maps an input graph to an Eulerian line-graph construction admitting unitary adjacency matrices, and (ii) a practical unitary projection procedure based on Newton-Schulz iteration. Theoretical analysis clarifies that, under standard analysis assumptions, unitary propagation keeps the graph-propagation term depth-stable, while vanilla normalized propagation exhibits exponential decay in its non-trivial spectral components. Across synthetic long-range tasks, TUDataset benchmarks, and LRGB datasets, GUMP improves over vanilla message passing and achieves competitive or superior performance against strong baselines. Code is available at \url{https://github.com/ucker/gump_code}.
\end{abstract}

\section{Introduction}
\label{sec:intro}

Unitarity has become an important principle for stabilizing deep neural networks and preventing gradient pathologies. Prior work has applied it through orthogonal or unitary initialization~\citep{saxe2013exact,arjovsky2016unitary,orvieto2023resurrecting,de2024griffin}, recurrent architectures with unitary transition matrices~\citep{arjovsky2016unitary,jing2017tunable}, and optimizers such as Muon~\citep{jordan2024muon,liu2025muon}. Across these settings, the common benefit is that unitary transformations preserve norms and make deep signal propagation easier to control.

\begin{center}
	\includegraphics[width=0.3\textwidth,trim=8 8 8 8,clip]{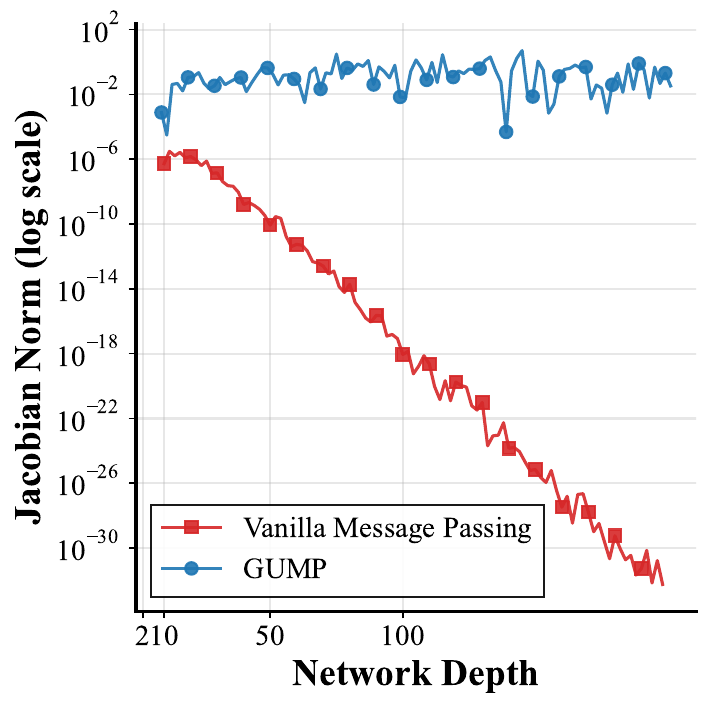}
	\hspace{0.02\textwidth}
	\includegraphics[width=0.3\textwidth,trim=8 8 8 8,clip]{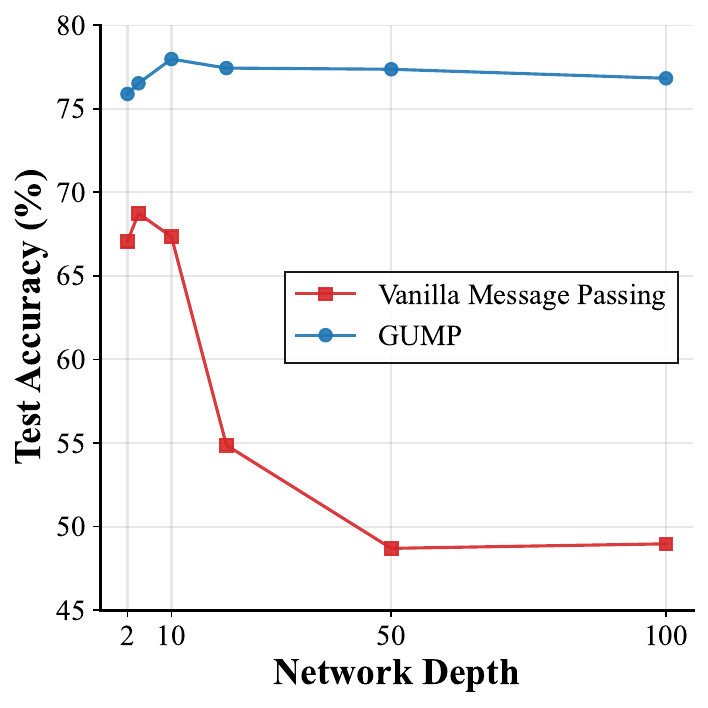}
    \vspace{-10px}
	\captionof{figure}{Compared to vanilla message passing (graph convolution), GUMP exhibits a stable Jacobian norm (left), which leads to more stable model performance as the number of layers increases (right).}\label{fig:intro}
    % \vspace{-10px}
\end{center}

However, constraining only the learnable parameters is insufficient for graph neural networks (GNNs), because repeated message passing is also governed by the graph structure itself. Nodes update their representations by aggregating information from neighbors~\citep{gilmer2017neural}, so the propagation dynamics depend on the adjacency operator rather than only on trainable weights. This creates graph-dependent forms of instability, most notably oversquashing and oversmoothing~\citep{alon2020bottleneck,topping2021understanding, chen2020measuring}, which limit the benefits of depth for long-range reasoning on graphs.

% However, imposing unitarity to parameters is not enough to address the learning efficiency problem of graph neural networks (GNNs) which arise from the graph structures through the message passing mechanism. GNNs rely on message passing mechanisms~\citep{gilmer2017neural} where each node aggregates feature vectors from its neighbors to update its representation. Unlike traditional neural networks where gradient flow is determined by learnable parameters, GNNs exhibit data-dependent learning inefficiency because the message passing is governed by the graph's adjacency structure. This inefficiency manifests as the oversquashing and oversmoothing problems~\citep{alon2020bottleneck,topping2021understanding, chen2020measuring}, which fundamentally limits GNNs' ability to stack layers for capturing long-range and complex dependencies in graphs.

Here, we extend the principle of unitarity from model parameters to the graph propagation operator itself. We propose Graph Unitary Message Passing (GUMP), a message-passing framework that replaces the usual normalized adjacency propagation with a unitary propagation operator on a transformed graph. The key point is not that all components of a GNN become unitary, but that the graph-induced factor in repeated propagation no longer decays exponentially with depth. This targets the graph-structure side of training instability, complementing standard practices that keep learnable weight matrices well conditioned.

% The key insight behind GUMP is that the measures of training stability in RNNs and GNNs share similar mathematical forms - both involve powers of transformation matrices that can lead to exponential gradient decay. Just as unitary RNNs solve the vanishing gradient problem by constraining the recurrent weight matrix to be unitary, GUMP alleviates oversquashing by imposing unitarity on the adjacency matrix used for message passing. This ensures that the Jacobian measure of oversquashing remains bounded rather than decaying exponentially with network depth.

To implement this idea, we transform a general graph into an Eulerian line-graph construction that admits unitary adjacency matrices while preserving the admissible edge-to-edge transitions induced by the original graph. We then use Newton-Schulz iteration~\citep{kovarik1970some,bjorck1971iterative} to compute the unitary propagation matrix efficiently. Experiments across diverse graph learning tasks show that GUMP improves over vanilla message passing and strong baselines.

% This superior performance stems from a key insight: achieving optimal results in graph neural networks requires tackling training instability from two angles. One must address not only traditional parameter-level instability but also the unique, data-dependent challenges inherent to graph structures. This dual approach is especially critical for ensuring the stability and performance of large-scale graph models.

\paragraph{Notations}
{We use bold uppercase letters for matrices, bold lowercase letters for vectors, and lowercase letters for scalars; for a matrix $\mathbf{X}$, $\mathbf{x}_i$ denotes row $i$ and $\mathbf{X}_{ij}$ an entry. The transpose and conjugate transpose are $\mathbf{X}^{\top}$ and $\mathbf{X}^{\dagger}$. Let $G=(V,E,\mathbf{X})$ be a graph with $n$ nodes, $e$ edges, and node features $\mathbf{X}\in\mathbb{R}^{n\times d}$; $\mathsf{V}[G]$ and $\mathsf{E}[G]$ denote its node and edge sets. We write $\tilde{\mathbf A}[G]$ for the binary adjacency, $\hat{\mathbf A}[G]=\mathbf{D}^{-1/2}\tilde{\mathbf A}[G]\mathbf{D}^{-1/2}$ for the normalized adjacency, and $\mathbf{A}[G]$ for a general matrix supported on $E$; when $G$ is clear, we omit the argument. The layer-$k$ GNN representation is $\mathbf{H}^{(k)}\in\mathbb{R}^{n\times d}$, with node vector $\mathbf{h}_i^{(k)}$. Additional graph-theory preliminaries are provided in the appendix.}

\section{Preliminaries on Unitarity and Training Instability}

\paragraph{Unitarity in Deep Learning}
{A matrix $\mathbf{U} \in \mathbb{C}^{n \times n}$ is \textit{unitary} if $\mathbf{U}^\dagger \mathbf{U}=\mathbf{U}\mathbf{U}^\dagger=\mathbf{I}$; for real matrices this is orthogonality. Its key property is norm preservation, $\|\mathbf{U}\mathbf{x}\|_2=\|\mathbf{x}\|_2$. This principle appears in orthogonal or unitary initialization~\citep{saxe2013exact,orvieto2023resurrecting}, unitary recurrent constraints for long-range sequence modeling~\citep{arjovsky2016unitary,jing2017tunable,orvieto2023resurrecting,de2024griffin}, and optimizers such as Muon~\citep{jordan2024muon,liu2025muon}. In these settings, controlling the spectrum or norm of the transformation helps stabilize deep signal propagation.}

\begin{figure*}[t]
	\centering
	\vspace{-20px}
	\includegraphics[width=1.0\textwidth]{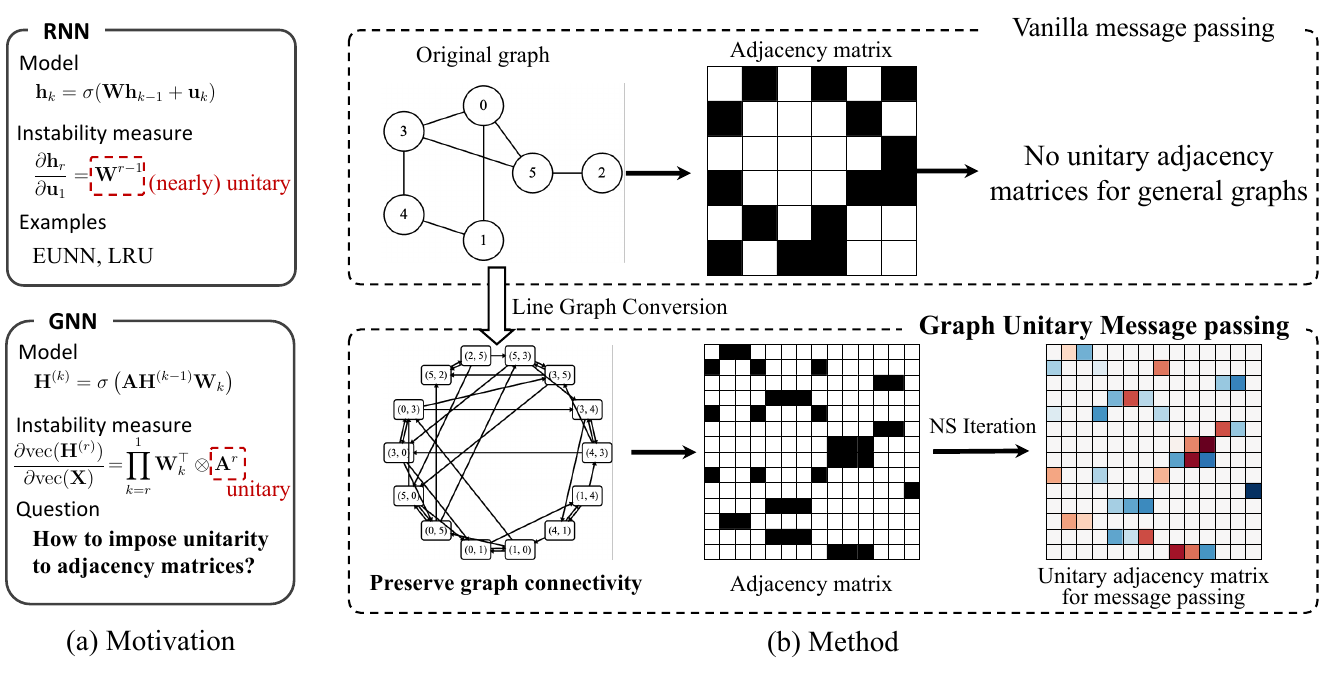}
	\vspace{-30px}
	\caption{Overview of GUMP.
		(a) RNN versus GNN. The measures of the stability are derived with the identity activation function and $\otimes$ denotes the Kronecker product.
		(b) GUMP aims to impose unitarity on $\mathbf{A}$ in message passing. GUMP achieves this goal with graph transformation (Section~\ref{ssec:transformation-unitary}) and unitary adjacency matrix calculation (Section~\ref{ssec:unitary-cal}).}
	\label{fig:gump}
	\vspace{-20pt}
\end{figure*}

\paragraph{Training Instability in GNNs}
To understand the training instability in GNNs, we first introduce the instability in RNNs from Figure~\ref{fig:gump}(a). We simply formulate RNN as $\mathbf h_k = \sigma(\mathbf W \mathbf h_{k-1} + \mathbf{u}_k)$ with $\mathbf h_k$ being the hidden state at layer $k$, $\mathbf W$ being the transformation matrix, $\mathbf{u}_k$ being the input at time step $k$, and $\sigma$ being the activation function.
The Jacobian of the hidden state with respect to the input is used to measure the training instability of RNN~\citep{arjovsky2016unitary}. If the Jacobian norm $\|\nicefrac{\partial \mathbf{h}_k}{\partial \mathbf{u}_0}\|_2$ grows exponentially with $k$, the gradient will explode, and if it decays exponentially, the gradient will vanish. When imposing unitarity on $\mathbf W$, the Jacobian norm is bounded, thus improves the training stability of RNNs.

% The success of imposing unitarity in RNNs for learning sequences motivates us to apply unitarity to GNNs to improve graph learning efficiency.

For graphs, the training instability is mainly caused by the message passing mechanism.
In Figure~\ref{fig:gump}(a), a one-hop MPNN layer for GNN is formulated as $\mathbf H^{(k)} = \sigma(\mathbf A \mathbf H^{(k-1)} \mathbf W_k)$ for simplicity. Similar to RNN~\citep{pascanu2013difficulty}, the Jacobian $\nicefrac{\partial \mathbf{H}^{(r)}}{\partial \mathbf{X}}$~\citep{topping2021understanding} can measure the training instability of GNN. When the Jacobian is large, the gradient will explode, and when it is small, the gradient will vanish. Therefore, we are motivated to impose unitarity on $\mathbf A$ in GNN to improve learning efficiency.

\section{Graph Unitary Message Passing}
\label{sec:motivate-GUMP}

% In this section, we propose Graph Unitary Message Passing (GUMP) to improve learning stability of GNNs. For simplicity, we consider the undirected graph in this paper. The extension to digraph is straightforward in appendix.
% \subsection{Framework of GUMP}

% \subsection{Transformation of graph to have unitary adjacency matrix}\label{ssec:transformation-unitary}

% \subsection{}\label{ssec:unitary-cal}

\subsection{Challenges of Imposing Unitarity}\label{ssec:motivation}

Imposing unitarity on an adjacency matrix is not as straightforward as that in RNNs.
The challenge comes from the sparsity of the adjacency matrix, as almost all unitary matrices are dense (Figure~\ref{fig:gump}(b)). Therefore, the number of graphs with unitary adjacency matrices is limited.
Also, the unitary adjacency matrix should be input-dependent and permutation-equivariant because the unitary adjacency matrix depends on input graph, and the order of nodes in a graph should not impact GNN representations.

In this section, the challenge of existence of unitary adjacency matrix is addressed by transforming the original graph to a special graph (Algorithm~\ref{alg:transformation}) which is guaranteed to have unitary adjacency matrices while preserving the admissible edge-to-edge transitions induced by the original graph at the same time.
The challenge of preserving adjacency matrix property is addressed by calculating the unitary adjacency matrix with a unitary projection algorithm (Algorithm~\ref{alg:efficient-GUMP}), which is implemented by utilizing the Newton-Schulz iteration and allows GUMP to be permutation-equivariant (Proposition~\ref{prop:permutation-equivariance}).
As a general one-hop message-passing mechanism, any convolution operation can be combined with GUMP by setting the edge weights to be the entries of the unitary adjacency matrix for message passing. The overview of GUMP is shown in Figure~\ref{fig:gump}(b).

\begin{algorithm}[h]
  % \footnotesize
    \caption{Graph transformation}
    \begin{algorithmic}
        \REQUIRE A undirected graph $G=(V,E)$;
        \STATE Initialize a new digraph $G^\prime = (V, E^\prime)$;
        \FOR{$(i,j)\in E$}
        \STATE Add $(i,j)$ and $(j,i)$ to $E^\prime$;
        \ENDFOR
        % \STATE remove duplicated edges in $E^\prime$;
        \STATE Convert $G^\prime$ to its line graph $\mathsf{L}(G^\prime)$;
        \STATE \textbf{Return:} A digraph $\mathsf{L}(G^\prime)$.
    \end{algorithmic}
    \label{alg:transformation}
\end{algorithm}

\subsection{Graph Transformation: Convert Graph to Have Unitary Adjacency Matrix}\label{ssec:transformation-unitary}

Since unitary matrices are generally non-symmetric, the graph transformation algorithm should convert the original graph to a directed graph.
We first formally define the unitary adjacency matrix as in \citet{severini2003digraph}.
Given an adjacency matrix $\mathbf{A}$, its support matrix $\textsf{S}[\mathbf{A}]\in \mathbb{R}^{n\times n}$ is a binary matrix with entries equal to one if the corresponding entry of $\mathbf{A}$ is non-zero and equal to zero otherwise, i.e.,
$\mathsf{S}[\mathbf{A}]_{ij}=1$, if $\mathbf{A}_{ij}\neq 0$ and $\mathsf{S}[\mathbf{A}]_{ij}=0$, if $\mathbf{A}_{ij}= 0$.
Then, the unitary adjacency matrix $\mathbf{U}_G$ of graph $G$ is a unitary matrix whose support is equal to the support of its adjacency matrix $\mathbf{A}$, i.e., $\mathsf{S}[\mathbf{U}_G] = \mathsf{S}[\mathbf{A}]$.
% \vspace{-10pt}

% \vspace{-15pt}

% \begin{algorithm}[]
% 	\caption{Graph transformation}
% 	\begin{algorithmic}[1]
% 		\REQUIRE A undirected graph $G=(V,E)$;
%         \STATE Initialize a new digraph $G^\prime = (V, E^\prime)$;
% 		\FOR{$(i,j)\in E$}
%         \STATE Add $(i,j)$ and $(j,i)$ to $E^\prime$;
% 		\ENDFOR
%         % \STATE remove duplicated edges in $E^\prime$;
%         \STATE Convert $G^\prime$ to its line graph $\mathsf{L}(G^\prime)$;
% 		\STATE \textbf{Return:} A digraph $\mathsf{L}(G^\prime)$.
% 	\end{algorithmic}
% \end{algorithm}

We propose the transformation in Algorithm~\ref{alg:transformation} for undirected graph $G$ to make it have unitary adjacency matrices while preserving the admissible edge-to-edge transitions induced by the original graph.
Algorithm~\ref{alg:transformation} first transforms the undirected graph into a digraph $G'$ by splitting each undirected edge into two directed edges. {For each connected component of $G$ that contains at least one edge, the corresponding component of $G'$ is connected and has equal in-degree and out-degree at every vertex, hence that component is Eulerian. In particular, when $G$ is connected, $G'$ is Eulerian.}
Then, it converts {the resulting digraph} to its line graph $\mathsf{L}(G')$ (see definition in appendix).
Finally, $\mathsf{L}(G')$ has unitary adjacency matrices because of its specularity and strong quadrangularity properties~\citep{severini2003digraph}, which is proved in the following proposition.

\begin{proposition}\label{prop:existence-alg}
The line graph $\mathsf{L}(G')$ returned by Algorithm~\ref{alg:transformation} has unitary adjacency matrices.
\end{proposition}
% \vspace{-10pt}
Proposition~\ref{prop:existence-alg} is proved in the appendix. {The proof treats disconnected inputs componentwise and combines the resulting unitary blocks by a block-diagonal direct sum.}
{Algorithm~\ref{alg:transformation} does not literally preserve node-level adjacency, because the vertices of $\mathsf{L}(G^\prime)$ correspond to directed edges of $G$. Rather, it preserves the admissible edge-to-edge transitions induced by the original graph: two vertices of $\mathsf{L}(G^\prime)$ are adjacent only when the corresponding directed edges in $G^\prime$ can be traversed consecutively through a shared endpoint. Thus, the construction does not introduce spurious transitions across disconnected components or unrelated parts of the original graph.}
Finally, Algorithm~\ref{alg:transformation} takes as input graph $G$ with $n$ nodes and $e$ edges, and outputs a line graph $\mathsf{L}(G^\prime)$ with $2e$ nodes.
{Its edge set contains exactly $\sum_{v\in V} d(v)^2$ directed edges in the undirected-to-Eulerian construction, because each vertex $v$ contributes all admissible transitions from its $d(v)$ incoming directed edges to its $d(v)$ outgoing directed edges. Therefore, constructing $\mathsf{L}(G^\prime)$ takes $\mathcal{O}\!\left(\sum_{v\in V} d(v)^2\right)$ time, which is $\mathcal{O}(|E|\Delta)$ for maximum degree $\Delta$ and $\mathcal{O}(|E|^2)$ in the worst case.}

% \vspace{-10px}
\subsection{Unitary Adjacency Matrix Calculation: Compute the Edge Weights for Message Passing}\label{ssec:unitary-cal}
% \section{Graph Unitary Message Passing}\label{sec:unitary-cal}

According to Proposition~\ref{prop:existence-alg}, the line graph $\mathsf{L}(G^\prime)$ has unitary adjacency matrices.
In this section, we propose an algorithm to calculate a unitary adjacency matrix for GUMP, because the unitary adjacency matrix depends on the input graph and should be calculated for each graph.

\subsubsection{Permutation-equivariant Projection}
Permutation equivariance of message passing is a key property for GNN to apply to graphs with varying node orders.
To achieve this, the calculation of unitary adjacency matrix has to be permutation equivariant.
GUMP consists of two steps: 1) calculate edge weights to form weighted adjacency matrix; 2) impose unitarity on weighted adjacency matrix.

Firstly, edge weight for $(i,j)\in \mathsf{E}[\mathsf{L}(G^\prime)]$ is calculated by
\begin{align}\label{eq:edge-weight}
\alpha_{ij} = \mathsf{Tanh}\left( {\mathbf{w}^\top \cdot \mathsf{LeakyReLU}(\mathbf{W}_s \mathbf{h}_i + \mathbf{W}_t \mathbf{h}_j)}  \right),
\end{align}
where $\mathbf h_i$ ($\mathbf h_j$) is the representations for node $i$ ($j$) in $\mathsf{L}(G^\prime)$,
$\mathbf{W}_s, \mathbf{W}_t \in \mathbb{R}^{d^\prime \times d}$ are transformation matrices for source and target nodes of an edge respectively, and $\mathbf{w}\in \mathbb{R}^{d^\prime}$ is a learnable parameter.
Then, the weighted adjacency matrix of $\mathsf{L}(G^\prime)$, denoted as $\mathbf{A}_w \in\mathbb{R}^{2e \times 2e}$, is formed from edge weights, i.e., $(\mathbf{A}_w)_{ij} = \alpha_{ij}$.

After calculating $\mathbf{A}_w$, we impose unitarity on $\mathbf{A}_w$ by projection.
We use the projection algorithm in \citet{keller1975closest}, which takes advantage of the fact that the polar transformation yields the closest unitary matrix to a given matrix in terms of the Frobenius norm, i.e., $\mathsf{U}[\mathbf{A}_w] = \arg\min_{\text{$\mathbf{U}$ is unitary}} \left\| \mathbf{A}_w-\mathbf{U} \right\|_F^2=\mathbf{A}_w\left( \mathbf{A}_w^\dagger \mathbf{A}_w\right)^{-\frac{1}{2}}$. The unitary projection $\mathsf{U}[\mathbf{A}_w]$ is guaranteed to be permutation-equivariant when $\mathbf{A}_w$ is a full-rank matrix with the following proposition.
% The following lemma describes the unitary projection in GUMP:
% \begin{lemma}\label{lemma:unitary-projection}
% Given a weighted adjacency matrix $\mathbf{A}_w\in\mathbb{R}^{2e \times 2e}$ of $\mathsf{L}(G^\prime)$, unitary projection of $\mathbf{A}_w$ is given by
% $\mathsf{U}[\mathbf{A}_w]\!\! :=\!\! \arg\min_{\text{$\mathbf{U}$ is unitary}} \left\| \mathbf{A}_w-\mathbf{U} \right\|_F^2 = \mathbf{A}_w
% \left( \mathbf{A}_w^\dagger \mathbf{A}_w\right)^{-\frac{1}{2}}$.
% \end{lemma}
% Lemma~\ref{lemma:unitary-projection} (proved in the appendix) indicates $\mathsf{U}[\mathbf{A}_w]=\mathbf{A}_w\left( \mathbf{A}_w^\dagger \mathbf{A}_w\right)^{-\frac{1}{2}}$ is the unitary adjacency matrix for GUMP.
% Also, the unitary projection $\mathsf{U}[\mathbf{A}_w]$ is guaranteed to be permutation-equivariant when $\mathbf{A}_w$ is a full-rank matrix with the following proposition.
\begin{proposition}[Strong permutation equivariance]\label{prop:permutation-equivariance}
    Given two permutation matrices $\mathbf{P}_1$ and $\mathbf{P}_2$, if $\mathbf{A}_w$ is a full-rank matrix, the unitary projection $\mathsf{U}[\mathbf{A}_w]$ is equivariant to both row and column permutations of $\mathbf{A}_w$, i.e., $\mathbf{P}_1 \mathsf{U}[\mathbf{A}_w] \mathbf{P}_2^\top = \mathsf{U}[\mathbf{P}_1 \mathbf{A}_w \mathbf{P}_2^\top]$.
\end{proposition}

By Proposition~\ref{prop:permutation-equivariance} (proved in appendix), the weighted adjacency matrix $\mathbf{A}_w$ should be full-rank to guarantee permutation equivariance of GUMP.
{Because the line-graph weighted adjacency decomposes into square transition blocks $\mathbf{B}_v$ as described below, this full-rank condition is equivalent to requiring every non-isolated block $\mathbf{B}_v$ to be full rank. Appendix~\ref{app:unitary-diagnostics} reports reviewer-requested diagnostics on the five TU datasets: using tolerance $10^{-7}$, all transition blocks are full rank on MUTAG, ENZYMES, NCI1, and NCI109, while PROTEINS has a 99.99\% full-rank rate. These results support the generic full-rank condition in the evaluated setting, but the formal equivariance statement still retains the full-rank assumption.}
However, the unitary projection
is computationally expensive due to the inverse square root of $\mathbf{A}_w^\dagger \mathbf{A}_w\in\mathbb{R}^{2e \times 2e}$.

{Before discussing the efficient implementation, we first clarify why the projection step is compatible with the sparse message-passing structure of $\mathsf{L}(G^\prime)$. Although polar factors are dense in general, the Eulerian line-graph construction gives $\mathbf{A}_w$ a special row/column block structure indexed by the intermediate vertex shared by two consecutive directed edges. The next proposition makes this structure explicit and shows that the projection preserves it.

\begin{proposition}[Admissible-transition support preservation]\label{prop:block-support}
Let $\mathbf{A}_{\mathsf L}$ and $\mathbf{A}_w$ be the binary and weighted adjacency matrices of $\mathsf{L}(G^\prime)$, where $G^\prime=(V,E^\prime)$ is returned by Algorithm~\ref{alg:transformation}. If rows are grouped by incoming directed edges $E_{\mathrm{in}}(v)$ and columns by outgoing directed edges $E_{\mathrm{out}}(v)$, then there exist permutation matrices $\mathbf{P}_{\mathrm{in}}$ and $\mathbf{P}_{\mathrm{out}}$ such that
\[
\mathbf{D}:=\mathbf{P}_{\mathrm{in}}\mathbf{A}_w\mathbf{P}_{\mathrm{out}}^\top
= \mathsf{diag}\!\big(\mathbf{B}_v\big)_{v\in V},
\]
where $\mathbf{B}_v$ contains the weights for transitions $E_{\mathrm{in}}(v)\to E_{\mathrm{out}}(v)$ and is square because $|E_{\mathrm{in}}(v)|=|E_{\mathrm{out}}(v)|$. Under the same row and column permutations, the binary line-graph adjacency satisfies
\[
\mathbf{P}_{\mathrm{in}}\mathbf{A}_{\mathsf L}\mathbf{P}_{\mathrm{out}}^\top
= \mathsf{diag}\!\big(\mathbf{J}_v\big)_{v\in V},
\qquad
\mathbf{J}_v=\mathbf{1}_{d(v)\times d(v)},
\]
where zero-size blocks for isolated vertices are omitted and each dense all-ones block $\mathbf{J}_v$ represents all admissible transitions $E_{\mathrm{in}}(v)\to E_{\mathrm{out}}(v)$. If $\mathbf{A}_w$ is full-rank, then
\[
\mathbf{P}_{\mathrm{in}}\mathsf{U}[\mathbf{A}_w]\mathbf{P}_{\mathrm{out}}^\top
= \mathsf{diag}\!\big(\mathsf{U}[\mathbf{B}_v]\big)_{v\in V}.
\]
Since each projected block $\mathsf{U}[\mathbf{B}_v]$ lies inside the corresponding dense support block $\mathbf{J}_v$, every nonzero entry of $\mathsf{U}[\mathbf{A}_w]$ lies on an admissible transition of $\mathsf{L}(G^\prime)$. Hence
$
\mathsf{S}[\mathsf{U}[\mathbf{A}_w]] \subseteq \mathsf{S}[\mathbf{A}_{\mathsf L}]
$
with equality whenever each block polar factor $\mathsf{U}[\mathbf{B}_v]$ is fully supported, which holds generically for continuous full-rank learned weights.
\end{proposition}

Proposition~\ref{prop:block-support} shows that GUMP does not rely on masking a dense polar factor after projection: the Eulerian line-graph construction induces the required block structure, and the polar projection preserves it.
}

\subsubsection{Efficient Implementation of Unitarity Projection}

Calculating the unitary projection $\mathsf{U}[\mathbf{A}_w]=\mathbf{A}_w (\mathbf{A}_w^\dagger \mathbf{A}_w)^{-\frac{1}{2}}$ directly involves computing the inverse square root of a matrix, which is computationally expensive and numerically unstable for large matrices, typically requiring singular value decomposition (SVD) with cubic complexity $\mathcal{O}((2e)^3)$. To address this, we employ Newton-Schulz iteration~\citep{kovarik1970some,bjorck1971iterative,jordan2024muon}, an iterative method that converges quadratically to the unitary polar factor without expensive decompositions.

The Newton-Schulz iteration~\citep{jordan2024muon} for computing the unitary factor $\mathbf{U}$ of a matrix $\mathbf{A}_w$ is given by the recurrence:
\begin{equation}
    \mathbf{X}_{t+1} = \frac{15}{8} \mathbf{X}_t - \frac{5}{4}\mathbf{X}_t \left( \mathbf{X}_t^\top \mathbf{X}_t \right) + \frac{3}{8} \mathbf{X}_t \left( \mathbf{X}_t^\top \mathbf{X}_t \right)^2, \quad \mathbf{X}_0 = \frac{\mathbf{A}_w}{\|\mathbf{A}_w\|_F}.
\end{equation}
Here, $\mathbf{X}_t$ converges to $\mathsf{U}[\mathbf{A}_w]$ as $t \to \infty$. The normalization by the Frobenius norm $\|\mathbf{A}_w\|_F$, which upper bounds the spectral norm, ensures the spectral norm is at most 1 for better convergence. Algorithm~\ref{alg:efficient-GUMP} uses a finite number of iterations and therefore returns an approximately unitary operator. {Appendix~\ref{app:unitary-diagnostics} reports the normalized residual $\|\mathbf{X}_{v,K}^{\top}\mathbf{X}_{v,K}-I\|_F/\sqrt{d_v}$ for $K=5$ and $K=10$. Increasing to $K=10$ drives the median residual to around $10^{-7}$ on all five TU datasets, while the larger p95/max values are confined to the tail of ill-conditioned blocks. This is the expected accuracy--cost trade-off of a fixed-iteration polar solver; because the computation is blockwise, the effect is localized and can be controlled by increasing $K$ or applying a more accurate block projection only to flagged near-singular blocks.} {Because $\mathbf{A}_w$ can be written as a block diagonal matrix after the row/column permutations in Proposition~\ref{prop:block-support},} the Newton-Schulz iteration can be applied efficiently either by sparse matrix multiplications or by applying the iteration separately to each block $\mathbf{B}_v$ of $\mathbf{D}$ after the corresponding row/column permutations, making it significantly more efficient than SVD-based approaches for the sparse but large adjacency matrices encountered in GNNs.
{This yields an explicit complexity bound for the actual algorithm. Since Algorithm~\ref{alg:transformation} replaces each undirected edge by both orientations, each block $\mathbf{B}_v$ has size $d(v)\times d(v)$, so every Newton--Schulz iterate decomposes into independent block updates. One iteration costs $\mathcal{O}\!\left(\sum_{v\in V} d(v)^3\right)$, and Algorithm~\ref{alg:efficient-GUMP} costs $\mathcal{O}\!\left(K\sum_{v\in V} d(v)^3\right)$ over $K$ iterations rather than $\mathcal{O}(K(2e)^3)$ for a naive dense implementation.}
{\paragraph{Complexity and memory.}
Let $M_L=\sum_{v\in V}d(v)^2$ be the number of admissible directed transitions in the line graph, $F$ the line-graph hidden dimension, and $F_a=d^\prime$ the attention dimension in \eqref{eq:edge-weight}. The graph transformation costs $\mathcal{O}(e+M_L)$ time and storage per input graph; learned edge weights cost $\mathcal{O}(2eFF_a+M_LF_a)$ time and $\mathcal{O}(M_L)$ storage after source/target projections are precomputed. The $K$-step blockwise Newton--Schulz projection costs $\mathcal{O}\!\left(K\sum_{v\in V}d(v)^3\right)$ time and $\mathcal{O}(M_L)$ forward storage, although automatic differentiation may retain $\mathcal{O}(KM_L)$ intermediates without checkpointing or a custom backward pass. A GCN-style GUMP propagation layer then costs $\mathcal{O}(M_LF+2eF^2)$, compared with $\mathcal{O}(eF+nF^2)$ for a standard GCN layer. Thus GUMP remains linear for bounded-degree sparse graphs but is degree-sensitive; high-degree vertices dominate $\sum_v d(v)^2$ and $\sum_v d(v)^3$.}
{\paragraph{Comparison with Kiani et al.}
\citet{kiani2024unitary} construct unitary graph convolutions on the original graph operator, for example through matrix-exponential unitary filters such as $\exp(i\mathbf{A})$. Unlike GUMP, they do not materialize a $2e$-node line graph or the $M_L=\sum_v d(v)^2$ edge-to-edge transition set, but their cost depends on how the unitary graph filters are constructed or approximated. Therefore, Kiani et al. avoids GUMP's line-graph expansion bottleneck, while its runtime depends on the construction or approximation of matrix-exponential unitary filters; a direct runtime comparison is implementation- and setup-dependent.}
% Motivated by Corollary~\ref{cor:same_support}, we apply a mask $\mathsf{S}[\mathbf{A}_w]$ to ensure that the resulting unitary adjacency matrix $\mathsf{U}[\mathbf{A}_w]$ retains the same sparsity pattern as $\mathbf{A}_w$ after the iterations.

\begin{algorithm}[h]
      % \footnotesize
    \caption{Blockwise calculation of unitary adjacency matrix via Newton-Schulz iteration}
    \begin{algorithmic}[1]
        \REQUIRE $\mathsf{L}(G^\prime)$ outputted by Algorithm~\ref{alg:transformation}, iterations $K$;
        \STATE Calculate $\mathbf{A}_w$ of $\mathsf{L}(G^\prime)$ with \eqref{eq:edge-weight};
        \STATE Find $\mathbf{P}_{\mathrm{in}}, \mathbf{P}_{\mathrm{out}}$ such that $\mathbf{D}=\mathbf{P}_{\mathrm{in}}\mathbf{A}_w\mathbf{P}_{\mathrm{out}}^\top=\mathsf{diag}(\mathbf{B}_v)_{v\in V}$ as in Proposition~\ref{prop:block-support};
        \FOR{each block $\mathbf{B}_v$}
            \STATE Normalize $\mathbf{X}_{v,0} = \mathbf{B}_v / \|\mathbf{B}_v\|_F$;
            \FOR{$t = 0$ to $K-1$}
                \STATE $\mathbf{X}_{v,t+1} = \frac{15}{8} \mathbf{X}_{v,t} - \frac{5}{4}\mathbf{X}_{v,t} \left( \mathbf{X}_{v,t}^\top \mathbf{X}_{v,t} \right) + \frac{3}{8} \mathbf{X}_{v,t} \left( \mathbf{X}_{v,t}^\top \mathbf{X}_{v,t} \right)^2$;
            \ENDFOR
        \ENDFOR
        \STATE $\mathsf{U}[\mathbf{D}] = \mathsf{diag}(\mathbf{X}_{v,K})_{v\in V}$;
        \STATE $\mathsf{U}[\mathbf{A}_w] = \mathbf{P}_{\mathrm{in}}^\top \mathsf{U}[\mathbf{D}] \mathbf{P}_{\mathrm{out}}$;
        \STATE \textbf{Return:} Approximately unitary adjacency matrix $\mathsf{U}[\mathbf{A}_w]$.
    \end{algorithmic}
    \label{alg:efficient-GUMP}
\end{algorithm}

\begin{algorithm}[h]
	% \footnotesize
	\caption{GNN with the graph unitary message passing mechanism (GNN-GUMP)}
	\begin{algorithmic}[1]
		\REQUIRE A graph $G=(V,E, \mathbf X)$;
        \STATE $\mathbf X^{(0)}=\mathsf{GNN}(\mathbf{X}, G)$;
        \STATE Transform $G$ to $\mathsf{L}(G^\prime)$ with Algorithm~\ref{alg:transformation};
        \STATE Generate initial representation $\mathbf{H}^{(0)}$ for $\mathsf{L}(G^\prime)$ with $\mathbf{h}_{(i,j)}\!=\! [\mathbf{x}^{(0)}_i ; \mathbf{x}^{(0)}_j ], \forall (i,j)\in \mathsf{V}[\mathsf{L}(G^\prime)]$;
        \STATE Calculate $\mathsf{U}[\mathbf{A}_w]$ with Algorithm~\ref{alg:efficient-GUMP};
        \FOR{$k = 1\cdots L$}
            % \STATE $\mathbf{H}^{(k)}=\sigma (\mathsf{U}[\mathbf{A}_w] \mathbf{H}^{(k-1)}\mathbf{W}_{k})$
            \STATE $\mathbf{h}_v^{(k)}=\gamma (\mathbf h_v^{(k-1)}, \sum_{u\in \mathcal{N}_{\mathsf{L}(G^\prime)}(v)}  \mathsf{U}[\mathbf{A}_w]_{vu} \phi^{(k)} (\mathbf{h}_v^{(k-1)}, \mathbf{h}_u^{(k-1)}))$, $v\in \mathsf{V}[\mathsf{L}(G^\prime)]$
        \ENDFOR
        \STATE Scatter $\mathbf{H}^{(L)}$ to nodes of $G$ with $\mathbf{H}_s^{(L)}\!\!=\! \mathsf{Scatter}(\mathbf{H}^{(L)}\!\!, G)$;
        \STATE Generate node representations of $G$ with $\mathbf{X}^{(L)}\!\!\! = \! [\mathbf{X}^{(0)};\mathbf{H}_s^{(L)}]$.
		\STATE \textbf{Return:} Node representations $\mathbf{X}^{(L)}$ of $G$.
	\end{algorithmic}
	\label{alg:GUMP}
\end{algorithm}

\subsection{Applying GUMP to GNNs}
\label{sec:gump}

\paragraph{Architecture and scatter operation.}
We can apply GUMP to different GNN architectures for graph learning tasks in Algorithm~\ref{alg:GUMP}.
Given a graph $G$, a base GNN first computes the initial node representations of $G$, i.e., $\mathbf{X}^{(0)}=\mathsf{GNN}(\mathbf{X}, G)$.
Then, Algorithm~\ref{alg:transformation} transforms $G$ to $\mathsf{L}(G^\prime)$.
The initial node representations $\mathbf H^{(0)}\in \mathbb{R}^{2e\times 2d}$ of $\mathsf{L}(G^\prime)$ are generated with 
$\mathbf{h}_{(i,j)} = [\mathbf{x}^{(0)}_i ; \mathbf{x}^{(0)}_j ]$,
$\forall (i,j)\in \mathsf{V}[\mathsf{L}(G^\prime)]$ ($i,j \in \mathsf{V}[G]$).
Next, the unitary adjacency matrix $\mathsf{U}[\mathbf{A}_w]$ of $\mathsf{L}(G^\prime)$ is calculated from Algorithm~\ref{alg:efficient-GUMP} and applied to propagate messages with
$\mathbf{h}_v^{(k)}=\gamma (\mathbf h_v^{(k-1)}, \sum_{u\in \mathcal{N}_{\mathsf{L}(G^\prime)}(v)}  \mathsf{U}[\mathbf{A}_w]_{vu} \phi^{(k)} (\mathbf{h}_v^{(k-1)}, \mathbf{h}_u^{(k-1)}))$, 
$v\in \mathsf{V}[\mathsf{L}(G^\prime)]$, where $\mathcal{N}_{\mathsf{L}(G^\prime)}(v)$ denotes the line-graph neighborhood of $v$. {Here, $\phi^{(k)}(\cdot,\cdot)$ denotes a generic layer-$k$ message function, and $\gamma(\cdot,\cdot)$ denotes the generic combine/update function that merges a node's previous representation with the aggregated message. In the method itself, GUMP constrains only the propagation operator $\mathsf{U}[\mathbf{A}_w]$; the learnable matrices inside the base GNN remain unconstrained. In the theory, we separately assume that the cumulative contribution of the weight matrices remains well conditioned, so that the analysis isolates the effect of repeated graph propagation.}
After $L$ layers of unitary message passing, we obtain the node representations $\mathbf{H}^{(L)}$, which is later scattered to nodes of $G$ with $\mathbf{H}_s^{(L)}=\mathsf{Scatter}(\mathbf{H}^{(L)}, G)\in\mathbb{R}^{n\times d^\prime}$.
Then, $\mathbf{H}_s^{(L)}$ are concatenated with $\mathbf{X}^{(0)}$ to obtain the final node representations $\mathbf{X}^{(L)} = [\mathbf{X}^{(0)}; \mathbf{H}_s^{(L)}]\in\mathbb{R}^{n\times (d+d^\prime)}$ of $G$.
Finally, various graph learning tasks, e.g., graph and node classification, link prediction, and graph regression, are performed based on $\mathbf{X}^{(L)}$.
In this paper, GUMP is a general one-hop message-passing mechanism for GNN. Therefore, depending on the convolution in line 6 of Algorithm~\ref{alg:GUMP}, GNN with GUMP is named as [GNN type]-GUMP in Section~\ref{sec:exps}, e.g., GCN-GUMP and GIN-GUMP have graph convolution and isomorphism operators in line 6 of Algorithm~\ref{alg:GUMP}, respectively.

{
Each node of $\mathsf{L}(G^\prime)$ corresponds to a directed edge $(u,v)$ in the bidirected graph $G^\prime$. We map directed-edge representations back to original nodes by aggregating the line-graph states whose directed edges terminate at the node. In our implementation, this scatter operation is a mean aggregation:
\[
\mathbf{h}^{(L)}_{s,v}
=
\frac{1}{|\mathcal{I}_{\mathrm{in}}(v)|}
\sum_{(u,v)\in \mathcal{I}_{\mathrm{in}}(v)}
\mathbf{h}^{(L)}_{(u,v)},
\]
where $\mathcal{I}_{\mathrm{in}}(v)$ denotes the set of incoming directed edges to $v$. For isolated nodes, which create no line-graph vertices, the scatter output is set to zero. We concatenate $\mathbf{H}_s^{(L)}$ with $\mathbf{X}^{(0)}$ as a residual node-level path. This preserves original node information that may not be represented by the edge-centric line graph, especially for isolated nodes that create no line-graph vertices. The concatenation also keeps local node-level features available to the task head while $\mathbf{H}_s^{(L)}$ contributes long-range unitary edge-to-edge propagation features.}

{
\paragraph{Expected-Jacobian motivation and row-energy stability.}
To connect the propagation analysis to training stability, consider the simplified message-passing dynamics
\[
\mathbf{H}^{(\ell)}=\mathsf{ReLU}(\mathbf{A}\mathbf{H}^{(\ell-1)}\mathbf{W}_{\ell}), \qquad \mathbf{H}^{(0)}=\mathbf{X}.
\]
The following factorization separates the cumulative feature-transformation term from the graph-propagation term.

\begin{proposition}[Expected-Jacobian factorization]
\label{prop:expected-jacobian-factorization}
Under the path-level random-gate approximation with a non-graph gate factor $\rho_L$, the expected Jacobian from source node $s$ to target node $i$ satisfies
\[
\mathbb{E}\!\left[
\frac{\partial \mathbf{h}_i^{(L)}}{\partial \mathbf{x}_s}
\right]
=
\rho_L
\left(\prod_{\ell=L}^{1}\mathbf{W}_{\ell}^{\top}\right)
(\mathbf{A}^{L})_{is}.
\]
\end{proposition}

Here $\rho_L$ is an analysis-only path-level factor summarizing ReLU gate effects along length-$L$ computation paths. It is treated as independent of the graph propagation operator. Proposition~\ref{prop:expected-jacobian-factorization} shows that, after separating the cumulative weight term and the non-graph gate factor, the graph-dependent depth behavior of the expected Jacobian is captured by the scalar propagation factor $(\mathbf{A}^{L})_{is}$.

For a fixed target node $i$, we therefore measure the source-averaged squared graph factor by
\[
I_i(L;\mathbf{A})
:=
\frac{1}{N}\sum_{s=1}^{N}|(\mathbf{A}^{L})_{is}|^{2}
=
\frac{1}{N}\|\mathbf{e}_i^\top \mathbf{A}^{L}\|_2^2.
\]
Equivalently,
\[
\frac{1}{N}\sum_{s=1}^{N}
\left\|
\mathbb{E}\!\left[
\frac{\partial \mathbf{h}_i^{(L)}}{\partial \mathbf{x}_s}
\right]
\right\|_F^2
=
|\rho_L|^2 I_i(L;\mathbf{A})
\left\|
\prod_{\ell=L}^{1}\mathbf{W}_{\ell}^{\top}
\right\|_F^2.
\]
Thus, $I_i(L;\mathbf{A})$ is not an additional architectural objective; it is the source-averaged graph factor appearing in the expected Jacobian.

Classical oversmoothing analyses often study the decay of non-stationary spectral components under repeated normalized propagation. In our setting, Proposition~\ref{prop:expected-jacobian-factorization} shows that the graph-dependent factor of the expected Jacobian is exactly $(\mathbf{A}^L)_{is}$. We therefore study its source-averaged squared magnitude, $I_i(L;\mathbf{A})$, which removes sign/phase cancellations and measures the row energy of the graph-propagation contribution. Theorem~\ref{thm:unitary-row-energy} shows that unitary propagation preserves this aggregate factor, while Theorem~\ref{thm:vanilla-spectral-decay} recovers the usual exponential decay of non-stationary modes for vanilla normalized propagation.

\begin{theorem}[Row-energy stability of unitary graph propagation]
\label{thm:unitary-row-energy}
    Let $\mathbf{A}\in\mathbb{C}^{N\times N}$ be unitary. For every target node $i\in[N]$ and depth $L\ge 0$,
    \begin{equation*}
        I_i(L;\mathbf{A})
        =
        \frac{1}{N}\|\mathbf{e}_i^\top\mathbf{A}^{L}\|_2^2
        =
        \frac{1}{N}.
    \end{equation*}
    Equivalently, unitary propagation preserves the source-averaged graph factor in the expected Jacobian. Hence, under the random-gate approximation and the well-conditioned-weight assumption, the graph propagation operator does not introduce exponential decay or explosion in this aggregate Jacobian measure.
\end{theorem}

Theorem~\ref{thm:unitary-row-energy} should be interpreted as a row-energy conservation result. Since $\mathbf{A}$ is unitary, $\mathbf{A}^L$ is also unitary for every depth $L$, so the squared $\ell_2$ norm of each row of $\mathbf{A}^L$ is preserved. Under the standard assumption that the cumulative weight factor $\prod_{\ell=L}^{1}\mathbf{W}_\ell^\top$ remains well conditioned and does not itself scale exponentially with $L$, any exponential decay or explosion in the expected Jacobian cannot come from the graph-propagation term. This is the intended claim of Theorem~\ref{thm:unitary-row-energy}.

The theorem does not claim that every individual entry $(\mathbf{A}^{L})_{is}$ is constant, uniformly lower bounded, or that the full Jacobian is depth-invariant. Individual source-target entries may increase, decrease, or oscillate because of sign or phase cancellations. The invariant is the row-energy aggregate $I_i(L;\mathbf{A})$, which is the source-averaged graph factor in the expected Jacobian.

\begin{theorem}[Vanilla spectral decay]
\label{thm:vanilla-spectral-decay}
For vanilla message passing with $\mathbf{A}=\hat{\mathbf{A}}\in\mathbb{R}^{N\times N}$, assume $\hat{\mathbf{A}}$ is real symmetric with spectral radius $1$ and eigenvalues
\[
1=\lambda_1>|\lambda_2|\ge\cdots\ge|\lambda_N|,
\qquad
c:=\max_{j\ge2}|\lambda_j|<1.
\]
Let $\{v_j\}_{j=1}^{N}$ be an orthonormal eigenbasis and $\mathbf{\Pi}:=v_1v_1^\top$. Then for every target node $i\in[N]$ and depth $L\ge0$,
\[
I_i(L;\hat{\mathbf{A}})
=
\mathbb{E}_{s\sim\mathrm{Unif}([N])}\left[\left|(\hat{\mathbf{A}}^L)_{is}\right|^2\right]
=
\frac{1}{N}v_{1,i}^2+
\frac{1}{N}\sum_{j=2}^N v_{j,i}^2\lambda_j^{2L},
\]
and therefore
\[
0\le
I_i(L;\hat{\mathbf{A}})-\frac{1}{N}v_{1,i}^2
\le
\frac{1}{N}c^{2L}.
\]
Equivalently, if $\delta_L(i,s):=(\hat{\mathbf{A}}^L-\mathbf{\Pi})_{is}$, then
\[
\mathbb{E}_s\left[\left|\delta_L(i,s)\right|^2\right]
=
\frac{1}{N}\sum_{j=2}^N v_{j,i}^2\lambda_j^{2L}
\le
\frac{1}{N}c^{2L}.
\]
Moreover, with $\alpha_i:=\sum_{j:\,|\lambda_j|=c}v_{j,i}^2$, we also have
$
\mathbb{E}_s\left[\left|\delta_L(i,s)\right|^2\right]
\ge
\frac{1}{N}\alpha_i c^{2L}$.
Thus, if $\alpha_i>0$, the non-stationary source-averaged influence decays as $\Theta(c^{2L}/N)$.
\end{theorem}

Theorem~\ref{thm:vanilla-spectral-decay} shows that vanilla normalized propagation preserves only the stationary component, while the non-stationary source-averaged graph influence decays exponentially with depth. In contrast, Theorem~\ref{thm:unitary-row-energy} shows that unitary propagation preserves $I_i(L;\mathbf{A})=1/N$, so GUMP removes the graph-induced exponential decay in this aggregate Jacobian factor. Proof is in Appendix C.5.
}

% \vspace{-0.5cm}

% \paragraph{Comparison with existing methods}
% GUMP paves a new way to improve training efficiency of graph learning.
% Overall, GUMP has the following advantages:
% (1) GUMP is permutation-equivariant, which is a desirable property for graph learning.
% (2) Unlike rewiring methods, GUMP does not introduce extra cross-component connectivity and instead preserves the admissible edge-to-edge transitions induced by the original graph.
% (3) GUMP achieves the optimal Jacobian measure of oversquashing since the eigenvalues of unitary adjacency matrices are complex units and thus will not change exponentially with respect to the number of GNN layers.

% Previous work on unitary GNN, e.g., Ortho-GConv~\citep{guo2022orthogonal},
% imposes unitarity on the feature transformation matrix of GNN.
% Unlike Ortho-GConv, GUMP addresses the issue of ill-posed gradient caused by oversquashing by imposing unitarity on the adjacency matrix.
% Moreover, enforcing unitarity on adjacency matrix is more challenging than that on feature transformation, 
% since adjacency matrices depend on the input graph and are not parameters of GNN. See the position of GUMP in Appendix~\ref{app:pos-of-gump}.

\FloatBarrier
\section{Experiments}\label{sec:exps}
In this section, we perform experiments to evaluate GUMP on graph learning tasks.
All experiments are implemented by PyTorch Geometric~\citep{fey2019pyg} and conducted on NVIDIA RTX 4090 GPUs and AMD EPYC 7763 CPUs.

% \vspace{-0.2cm}

\subsection{Experiments on Synthetic Dataset}

\begin{figure}[t]
    \vspace{-5px}
    \centering
    \subfigure[CrossedRing]{
		\includegraphics[width=0.25\textwidth]{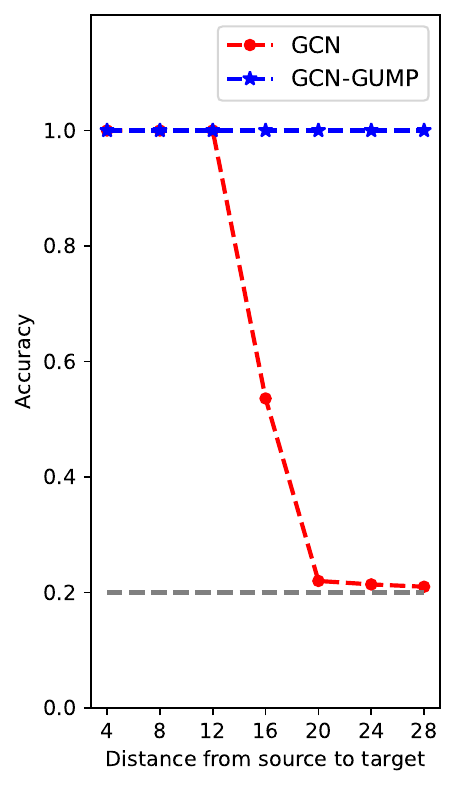}
		\label{fig:cross_ring}}
    \hspace{-3px}
    \subfigure[Ring]{
		\includegraphics[width=0.25\textwidth]{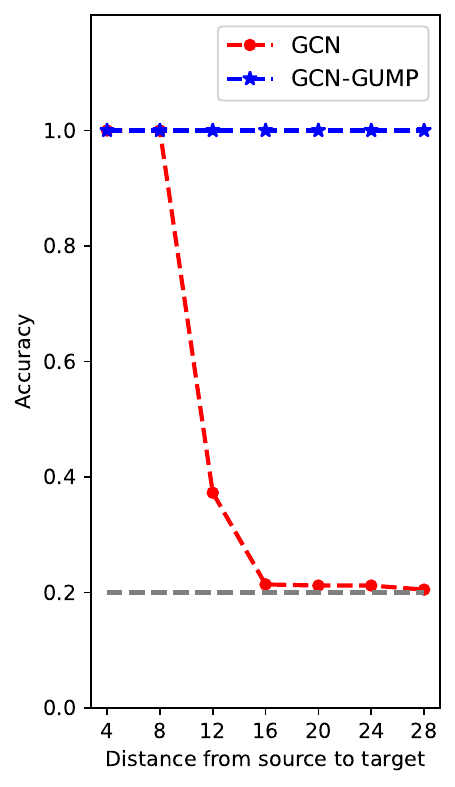}
		\label{fig:ring}}
    \hspace{-3px}
    \subfigure[CliquePath]{
		\includegraphics[width=0.25\textwidth]{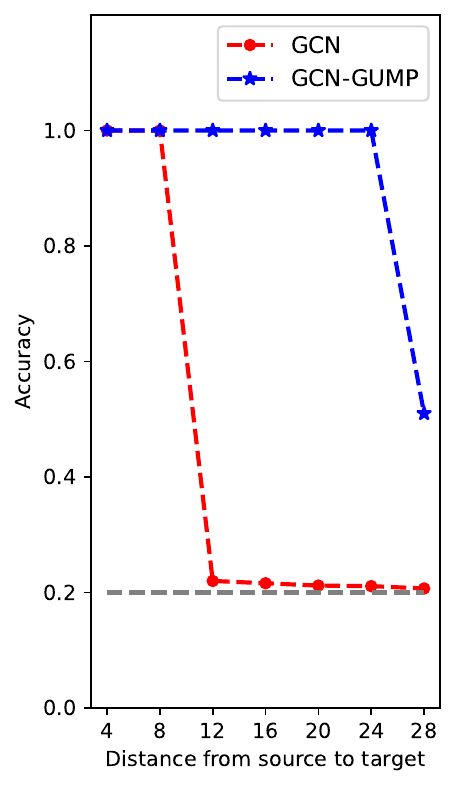}\label{fig:lollipop}}
	\vspace{-10px}
    \caption{The performance of GCN and GCN-GUMP on the CrossedRing, Ring, and CliquePath with different distances from source to target.}\label{fig:synthetic}
    \vspace{-10px}
\end{figure}

\paragraph{Setup} In this section, we conduct experiments on synthetic datasets, i.e., CrossedRing, Ring, and CliquePath, 
in \citet{di2023over} to test GUMP.
The performance is evaluated on the distances from source to target in the range of 4 to 28.
This experiment is designed to test the ability of GUMP to learn long-range interactions when increasing the model layers.
In the experiments, we compare GCN-GUMP and GCN.
The layer $L$ of GCN-GUMP and GCN is appropriately set up according to the distance $d$ between source and target in the synthetic datasets (i.e., $L= \lfloor d/2\rfloor+1$), such that the long-range interactions can be captured by GNN.
We set the hidden dimension to be $32$ for both GCN-GUMP and GCN.
The hyperparameters of GCN-GUMP for synthetic datasets are in appendix.

% \vspace{-0.2cm}

\paragraph{Results} We plot the average results from three random seeds of GCN-GUMP and GCN experiments in Figure~\ref{fig:synthetic}. For two easier datasets, i.e., CrossedRing and Ring, GUMP achieves 100\% accuracy when the distance ranges from 4 to 28.
For the challenging CliquePath dataset, GCN-GUMP's performance deteriorates to random guessing at a distance of 28.
The results show that GUMP can help capture the long-range interactions in graph learning tasks. We compare with more baselines in appendix.

% \begin{wrapfigure}{r}{0.6\textwidth}

% \end{wrapfigure}
\vspace{-0.2cm}
\subsection{Experiments on the TUDataset}

\begin{table*}[t]
   \centering
  %  \scriptsize
   \caption{Graph classification accuracy on the TUDataset. {\bf First} and {\underline{second}} best results are bold and underlined, respectively.}\label{tab:graph-classification-long-range}
   \setlength\tabcolsep{4pt}
   \begin{tabular}{c|c|ccccc}
   \toprule
           Classes & Methods & Mutag       & Proteins     & Enzymes       & NCI1 & NCI109  \\
            \midrule
            \multirow{2}{*}{--} & GIN     & 77.70±3.60 & 70.80±0.83  & 33.80±1.12   & 75.65±0.49 & 74.93±0.46 \\
   & GIN (+layer)         & 69.80±2.75  & 68.71±0.96   &   25.92±1.07     & 73.49±0.46 & 72.47±0.53   \\ \midrule
   % Last FA  & 70.05±2.03 & 71.02±0.96  & 26.47±1.20  & 48.98±0.95  \\
   % Every FA & 70.45±1.96 & 60.04±0.93  & 18.33±1.04  & 48.17±0.80  \\
   \multirow{5}{*}{\makecell{Rewiring\\(GIN)}} & DIGL     & {79.80±2.08} & 70.71±0.67  & 35.74±1.20   & \underline{79.37±0.43} & 76.88±0.39  \\
    & SDRF     & 78.40±2.80 & 69.81±0.79  & {35.82±1.09}    & 74.55±0.54 & 73.89±0.43  \\
    & FoSR     & 78.00±2.22 & \underline{75.11±0.82}  & 29.20±1.38  & 70.15±0.47& 69.93±0.45  \\
    & GTR      & 77.60±2.84 & {73.13±0.69}  & 30.57±1.42   & {75.45±0.44} & {75.28±0.42}  \\ 
    & ELDANADD &  \underline{82.16±0.03} & 70.53±0.86 & 26.36±0.01 & -- & -- \\ \midrule
   \multirow{2}{*}{Diffusion} &ADGN &	81.39±1.81	& 73.81±0.80	& 28.78±1.25	& 76.15±0.42	& 74.31±0.44 \\
   & GRAND	& 77.94±1.73	& 73.24±0.94	& 24.13±1.05	& 68.51±0.48	& 67.26±0.46 \\ \midrule
   \multirow{1}{*}{Transformer} & GPS	& 70.12±1.68 &	69.32±0.86	& 30.43±1.76	& 61.32±0.51	& 61.01±0.22 \\ \midrule
   \multirow{1}{*}{--} & Ortho-GConv	& 71.78±2.52	& 63.80±0.98	& 18.30±1.13	& 69.92±0.60	& 68.91±0.50 \\ \midrule
   \multirow{1}{*}{--} & Kiani et al. & 78.00±1.96 & 71.57±0.60 & \underline{43.67±1.45} & 78.71±0.42 & \underline{78.10±0.34} \\ \midrule
   \multirow{1}{*}{K-hop MP} & GRIT & 80.76±2.18 & 73.71±0.89 & 35.22±1.17 & 72.21±0.46 & 71.68±0.44 \\ \midrule
   \multirow{1}{*}{Ours} & GIN-GUMP     & {\bf 87.11±7.61} & {\bf 77.53±2.34} &  {\bf 53.11±5.80}     & {\bf 80.78±1.57} & {\bf 80.50±1.63}  \\
    \bottomrule
   \end{tabular}
   \vspace{-10px}
\end{table*}

\paragraph{Datasets}
We select five TUDataset benchmarks~\citep{morris2020tudataset}: Mutag, Proteins, Enzymes, NCI1, and NCI109. These chemistry or biological graphs contain complex interactions, including cases where atoms far apart in graph distance may be close in space. Dataset statistics are in the appendix.

% \vspace{-0.3cm}

\paragraph{Baselines}
Baselines include various rewiring methods, i.e., DIGL~\citep{gasteiger2019diffusion}, SDRF~\citep{topping2021understanding}, FoSR~\citep{Karhadkar2023}, GTR~\citep{black2023understanding}, and ELDANADD~\citep{jamadandi2024spectral}, diffusion methods, i.e., ADGN~\citep{tonshoff2023did} and GRAND~\citep{chamberlain2021grand}, transformers method GPS~\citep{rampavsek2022recipe}, GNN with orthogonal parameters Ortho-GConv~\citep{guo2022orthogonal}, k-hop message passing GRIT~\citep{di2023over}, and Kiani et al.~\citep{kiani2024unitary}.
The baselines' settings follow \citet{Karhadkar2023}.
{We refer to the unitary-convolution baseline of \citet{kiani2024unitary} as ``Kiani et al.'' in all tables for consistency.}
For the Kiani et al. entries on these five TU datasets, we reproduce the results by running the released code of \citet{kiani2024unitary} under the TU evaluation protocol.

% \vspace{-0.3cm}

% Since GNNs for these baselines have four layers, we also increase the number of layers of baselines to eight to show the performance of baselines with more layers (None(+layer)).

\paragraph{Experimental Details} 
For each method, 10\% of graphs are held out for testing and the remaining 90\% form the development set. We use 100 random train/validation splits of the development set, with 80\% for training and 10\% for validation, stop with patience 100 based on validation loss, and report test accuracy with 95\% confidence intervals for the best validation setting across runs. GUMP depth is manually tuned because capturing long-range interactions requires sufficient layers; detailed hyperparameters are in the appendix.

% The number of layers for rewiring methods is set to be in the range of one to five.

% In experiments, we set the base GNN to be GCN~\citep{kipf2016semi} and GIN~\citep{xu2018powerful} with one layer.
\begin{figure*}[t]
    \centering
    \vspace{-5px}
    \subfigure[Jacobian norm versus layers on NCI1]{
		\includegraphics[width=0.32\textwidth]{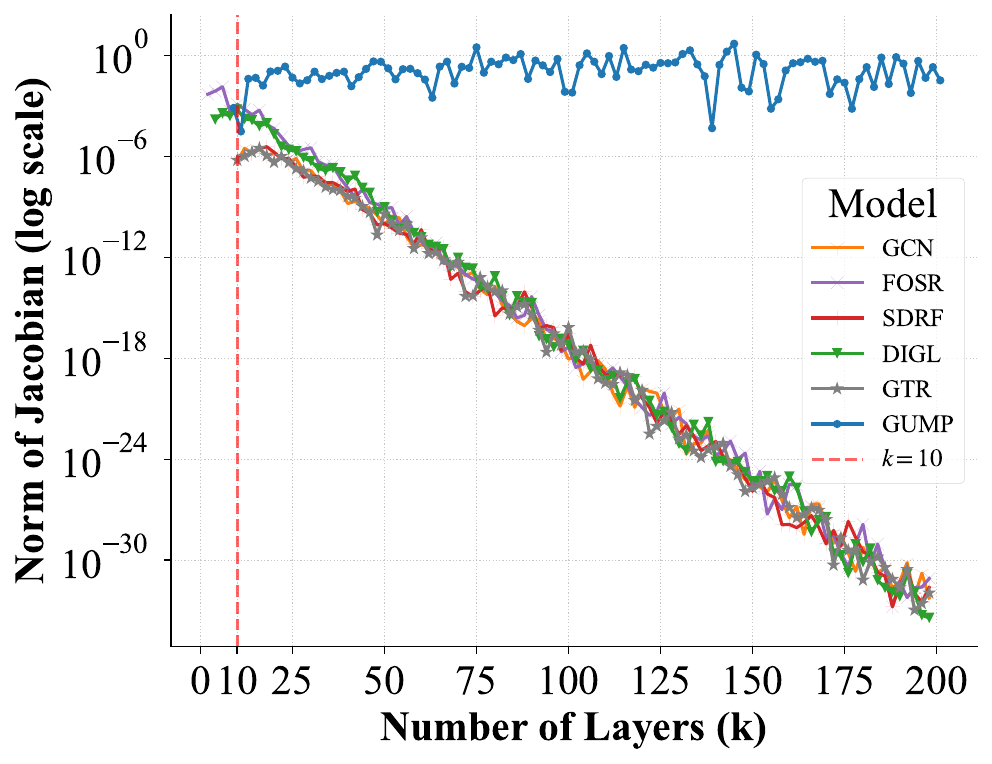}
		\label{fig:jac-vis}
	}
    \subfigure[Accuracy versus layers on NCI1]{
		\includegraphics[width=0.31\textwidth]{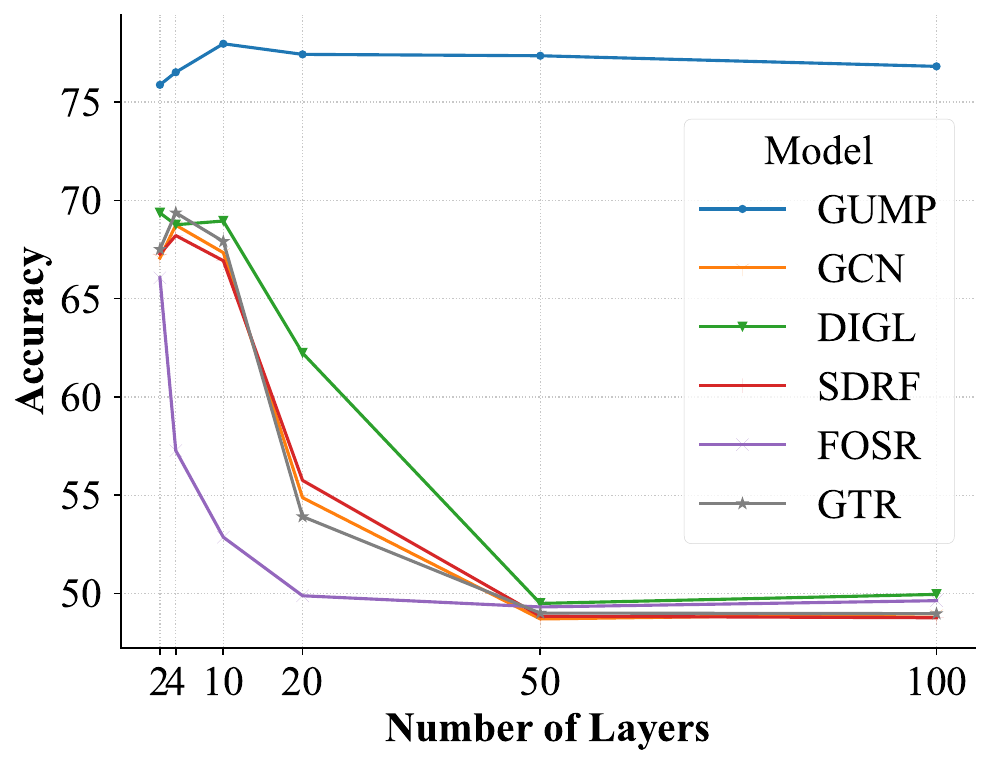}
		\label{fig:acc-vs-layer}
	}
    \subfigure[Ablation on different components]{
		\includegraphics[width=0.31\textwidth]{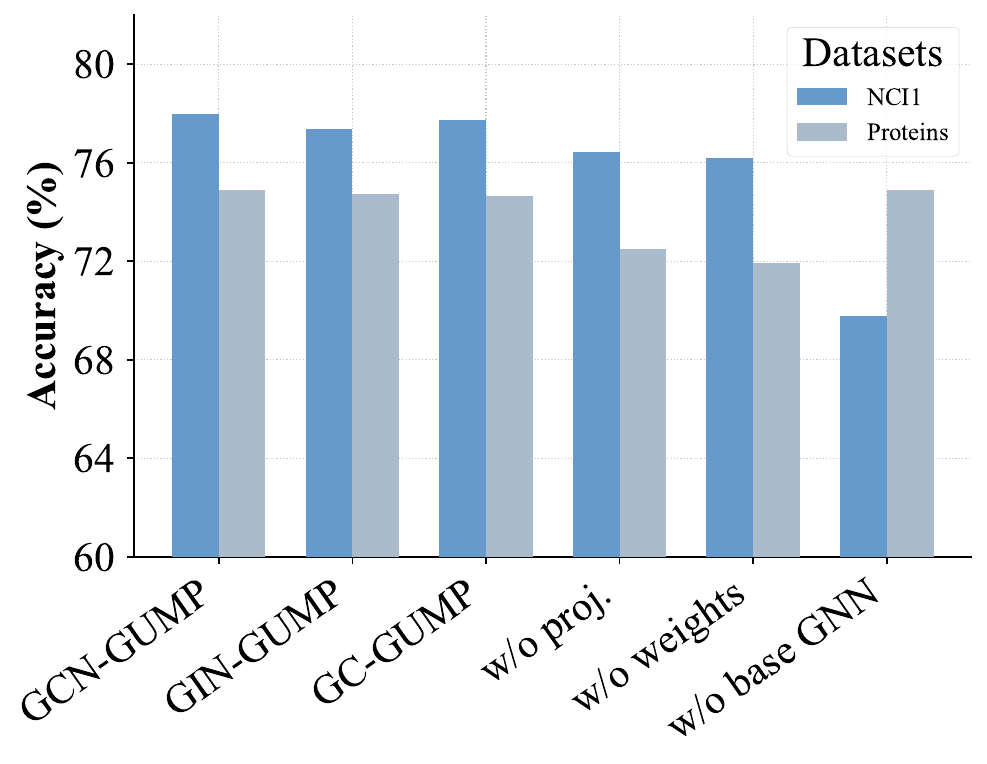}\label{fig:ablation-components}
	}
    % \includegraphics[width=0.45\columnwidth]{img/jac_measure.pdf}
    % \missingfigure[figwidth=\columnwidth]{GUMP architecture}
    % \vspace{-15px}
    \caption{Model analysis. GCN, GIN, and GUMP in (c) represent the convolution of GUMP. The base GNN of GUMP is GCN. ``w/o proj'' removes unitary projection in GUMP. ``w/o weights'' removes weighted adjacency matrix and unitary projection in GUMP. ``w/o base GNN'' removes base GNN in GUMP.}
    % \vspace{-10px}
\end{figure*}

% \vspace{-0.2cm}

\paragraph{Results}
The results of GUMP on TUDataset are shown in
% \footnote{+qm+ duplicate table labels.}
Table~\ref{tab:graph-classification-long-range}.
GUMP achieves superior performance across all five datasets, with GIN-GUMP obtaining the best results, demonstrating that the unitary message passing mechanism effectively learns complex interactions in graphs.
Moreover, since GUMP usually has more layers than baselines in these datasets, experiments show that GCNs with more layers exhibit degraded performance on all datasets, showing that improvement of GUMP does not come from increasing expressivity with more GNN layers.

\subsection{Experiments on LRGB}

\begin{wraptable}{r}{0.62\textwidth}
  \centering
%   \vspace{-20px}
  \scriptsize
  \setlength\tabcolsep{2pt}
  \caption{Results of Peptides-func and Peptides-struct. \textbf{Bold} and \underline{underline} denote best and second-best results.}\label{tab:graph-classification-lrgb}
  \begin{tabular}{r|cc}
    \toprule
    Methods & Peptides-func  & Peptides-struct \\
    & Test AP $\uparrow$ & Test MAE $\downarrow$ \\ \midrule
    GCN & 0.6860±0.0050 & {0.2460±0.0007} \\
    SDRF & 0.6874±0.0032 & 0.2453±0.0018 \\
    FoSR & 0.6878±0.0030 & 0.2461±0.0016 \\
    GTR & 0.6740±0.0033 & 0.2509±0.0013 \\
    LASER & 0.6440±0.0010 & 0.3043±0.0019 \\
    GRAND & 0.5789±0.0062 & 0.3418±0.0015 \\
    ADGN & 0.5975±0.0044 & 0.2874±0.0021 \\
    GPS & 0.6534±0.0091 & 0.2509±0.0014 \\
    Exphormer & 0.6527±0.0043 & 0.2481±0.0007 \\
    GRIT & 0.6988±0.0082 & 0.2460±0.0012 \\
    Graph ViT & 0.6942±0.0075 & 0.2449±0.0016 \\
    G-MLPMixer & 0.6921±0.0054 & 0.2475±0.0015 \\
    Drew-GCN & 0.6996±0.0076 & 0.2781±0.0028 \\
    ProxyAdd+GCN & 0.6789±0.0002 & 0.2465±0.0004 \\
    k-GCN-SSM & 0.6902±0.0138 & 0.2898±0.0324 \\
    Kiani et al. & \underline{0.7072±0.0035} & \textbf{0.2425±0.0009} \\
    GUMP & \textbf{0.7088±0.0026} & \underline{0.2438±0.0014} \\ \bottomrule
  \end{tabular}
%   \vspace{-10px}
\end{wraptable}

We evaluate on the Peptides-func and Peptides-struct datasets from the Long Range Graph Benchmark (LRGB)~\citep{dwivedi2022long}; dataset statistics are in the appendix.
Following \citet{tonshoff2023did}, we compare GUMP with reproduced long-range graph baselines, including SDRF, FoSR, GTR, LASER~\citep{barbero2023locality}, GRAND, ADGN, GPS, Exphormer~\citep{shirzad2023exphormer}, GRIT, Graph ViT and G-MLPMixer~\citep{he2023generalization}, Drew-GCN~\citep{gutteridge2023drew}, ProxyAdd+GCN~\citep{jamadandi2024spectral}, and k-GCN-SSM~\citep{arroyo2025vanishing}. For GUMP, the base one-hop operator is GCN; the appendix reports all tuned layers, optimization settings, and training hyperparameters. We also include the Kiani et al. results reported by \citet{kiani2024unitary} as a recent unitary-convolution baseline, noting that those numbers follow that paper's setup rather than our reproduced protocol.
% The base GNNs for GUMP and all baselines are set to GCN.
% The base GCN layer of GUMP is set to three, and the layer of GUMP is set to be eight in these three datasets.
The hyperparameters of GUMP and more results for LRGB are presented in appendix.
% All datasets are tested without any additional features, e.g., positional encoding.

The results of LRGB are shown in Table~\ref{tab:graph-classification-lrgb}. GUMP achieves the best result on Peptides-func and the second-best result on Peptides-struct, improving over GCN on both datasets while remaining slightly behind the reported Kiani et al. result on Peptides-struct. Overall, GUMP is competitive with recent unitary-convolution methods on LRGB while using a sparse, input-dependent unitary message-passing construction.

% \usepackage{multirow}
% \begin{table}[]
%     \caption{Results of \texttt{Peptides-func}, \texttt{Peptides-struct}, and \texttt{PCQM-contact} with GCN. Best results are \textbf{bold}.}
%     \label{tab:graph-classification-lrgb}
%     \centering
%      \setlength\tabcolsep{3pt}
% %    \setlength\tabcolsep{5pt}
%     \begin{tabular}{c|ccc}
%     \toprule
%     \multirow{2}{*}{} & Peptides-func  & Peptides-struct & PCQM-contact  \\
%                       & Test AP $\uparrow$       & Test MAE $\downarrow$       & Test MRR $\uparrow$     \\ \midrule
%     None              & .5930±.0023 & .3496±.0013   & .3234±.0006 \\
%     % None(+layer)              & .5930±.0023 & .3496±.0013   & .3234±.0006 \\
%     SDRF              & .5947±.0035 & .3404±.0015   & .3249±.0006 \\
%     FoSR              & .5947±.0027 & .3078±.0026   & .2783±.0008 \\
%     GTR               & .5075±.0029 & .3618±.0010   & .3007±.0022 \\
%     % BORF              & .6012±.0031 & .3374±.0011   & TIMEOUT       \\
%     LASER             & .6440±.0010 & .3043±.0019   & .3275±.0011 \\
%     GUMP              & \bf.6843±.0037 & \bf.2564±.0023   &   \bf .3413±.0015  \\ \bottomrule          
%     \end{tabular}
% \end{table}

% \vspace{-0.2cm}

\subsection{Model Analysis}\label{ssec:exp-analysis}
We analyze GUMP through Jacobian norms, depth, and ablations.

% \begin{wraptable}{r}{0.55\textwidth}  
% 	\centering
% 	\vspace{-10pt}
% 	\begin{tabular}{r|cc}
% 		\toprule
% 		Methods & Peptides-func  & Peptides-struct \\
% 		& Test AP $\uparrow$ & Test MAE $\downarrow$ \\ \midrule
% 		GCN & 0.6860±0.0050 & \textit{0.2460±0.0007} \\
% 		SDRF & 0.6874±0.0032 & 0.2453±0.0018 \\
% 		FoSR & 0.6878±0.0030 & 0.2461±0.0016 \\
% 		GTR & 0.6740±0.0033 & 0.2509±0.0013 \\
% 		LASER & 0.6440±0.0010 & 0.3043±0.0019 \\
% 		GRAND & 0.5789±0.0062 & 0.3418±0.0015 \\
% 		ADGN & 0.5975±0.0044 & 0.2874±0.0021 \\
% 		GPS & 0.6534±0.0091 & 0.2509±0.0014 \\
% 		Exphormer & 0.6527±0.0043 & 0.2481±0.0007 \\
% 		GRIT & 0.6988±0.0082 & 0.2460±0.0012 \\
% 		Graph ViT & 0.6942±0.0075 & 0.2449±0.0016 \\
% 		G-MLPMixer & 0.6921±0.0054 & 0.2475±0.0015 \\
% 		GUMP & \textbf{0.7088±0.0026} & \textbf{0.2438±0.0014} \\ \bottomrule
% 	\end{tabular}
% 	\caption{Results of Peptides-func and Peptides-struct. \textbf{Bold} are best results.}\label{tab:graph-classification-lrgb}
% \end{wraptable}

% In this section, we perform more experiments to analyze the GUMP and baselines.
\paragraph{Jacobian} {We visualize the spectral norm of the Jacobian for a distance-ten node pair from NCI1, using GCN as the base GNN. Figure~\ref{fig:jac-vis} shows that GUMP remains stable as depth increases, while the baselines decay rapidly. Although this operator-norm diagnostic differs from the row-energy quantity $I_i(L;\mathbf{A})$, both probe repeated graph propagation: normalized propagation suppresses non-stationary modes, whereas unitary propagation preserves their magnitudes. The trend is therefore directionally consistent with Theorems~\ref{thm:unitary-row-energy} and~\ref{thm:vanilla-spectral-decay}. DIGL has a larger Jacobian norm than other rewiring methods for depths below 50, matching its stronger performance in Table~\ref{tab:graph-classification-long-range}.}

% \vspace{-0.2cm}

\paragraph{Deep GNN} We increase the number of layers of GUMP and baselines to test depth robustness.
% \footnote{+qm+ why another dataset is used?
% 	why not make it 200 layers?}
This experiment is conducted in NCI1 with base GNN as GCN and the number of layers in the range of 2 to 100.
% The other hyperparameters of GUMP are the same as Table~\ref{tab:hyperparameters}.
% \todo{update res}
The results in Figure~\ref{fig:acc-vs-layer} demonstrate that the performance of GUMP increases from 75.88\% to 77.97\% when increasing the number of layers from 2 to 10.
However, baseline performance decreases substantially with depth.
For example, the performance of FoSR decreases from 66.06\% to 52.86\% when the number of layers increases from 2 to 10.
The baselines change drastically as layers increase, while GUMP is more stable, indicating that it can use deeper propagation for long-range interactions.

\vspace{-0.2cm}

\paragraph{Ablation Studies} Figure~\ref{fig:ablation-components} analyzes GUMP components. Replacing the GUMP convolution with GIN or GraphConv~\citep{morris2019weisfeiler} changes performance, showing sensitivity to the underlying convolution. Removing the unitary projection (message passing with $\mathbf{A}_w$) or both weights and projection (message passing with $\tilde{\mathbf{A}}[\mathsf{L}(G')]$) degrades performance, supporting their importance. Removing the base GNN decreases performance on NCI1 but not on Proteins, consistent with the dependence of $\mathbf{A}_w$ on line-graph node representations in \eqref{eq:edge-weight}.
% \todo{add for removing unitary projection}

Overall, GUMP shows the most depth-stable Jacobian behavior among the compared methods. {Appendix~\ref{app:exp} reports dataset statistics, hyperparameters, additional comparisons, and Jacobian visualizations; Appendix~\ref{app:node-classification} reports node-classification and runtime results, including Table~\ref{tab:runtime}, which motivates the scalability limitation in Section~\ref{sec:limitations}.}

\section{Discussions and Limitations}\label{sec:limitations}

In this paper, we propose Graph Unitary Message Passing (GUMP), which propagates messages using unitary adjacency matrices and performs well across several graph learning tasks. We discuss below its limitations and future work.

{\paragraph{Scalability to large graphs.}
The current implementation is less suitable for large single-graph settings and high-degree vertices. The line-graph transformation expands an input with $e$ undirected edges into $2e$ directed-edge states and $M_L=\sum_{v\in V}d(v)^2$ transitions; message passing stores $\mathcal{O}(2eF)$ states and processes $\mathcal{O}(M_LF)$ messages per layer, while blockwise projection costs $\mathcal{O}(K\sum_v d(v)^3)$. Appendix~\ref{app:node-classification}, Table~\ref{tab:runtime} shows that GUMP is slower than GCN under a matched propagated-node mini-batch budget, so GUMP is most appropriate when depth stability or long-range propagation justifies this extra cost, especially on small- and medium-scale graph classification or bounded-degree sparse graphs. Reducing the projection's dependence on high-degree vertices remains important for scaling GUMP. Scaling will likely require line-graph sampling, degree-aware transition sparsification, or approximate block-unitary propagation.}

{\paragraph{Finite-step Newton--Schulz approximation.}
Our stability analysis uses the exact polar factor $\mathsf{U}[\mathbf{A}_w]$, while Algorithm~\ref{alg:efficient-GUMP} returns an approximately unitary operator after $K$ Newton--Schulz iterations. The finite iterates preserve admissible block support and improve conditioning, but the residuals in Appendix~\ref{app:unitary-diagnostics} show an accuracy--cost trade-off on ill-conditioned blocks. The larger residuals are concentrated in the ill-conditioned tail under a fixed iteration budget, so they reflect numerical accuracy rather than a structural failure of the construction. Because projection is blockwise, this numerical effect is local and can be controlled by increasing $K$ or selectively applying a higher-accuracy projection to near-singular blocks. A complete theory should quantify how residual non-unitarity accumulates with depth and depends on block condition numbers and graph degrees.}

\paragraph{Information Loss}
For undirected, unweighted graphs, GUMP's transformation is lossless. For weighted or directed graphs, edge features encode weights or directions following R-GNNs~\citep{battaglia2018relational}, and post-processing filters invalid directed edges. {These adaptations preserve the essential graph information used by GUMP while acknowledging that weighted and directed settings require explicit edge attributes and task-specific filtering.}

% \paragraph{Computational Complexity}
% GUMP incurs computational overhead from unitary projection construction, particularly for large, dense graphs. While this overhead is essential for optimal Jacobian stability (Theorem~\ref{thm:unitary-row-energy}) and superior performance (Figure~\ref{fig:ablation-components}), GUMP is best suited for accuracy-critical applications with moderate-sized, sparse graphs. Future work will explore diagonalized adjacency spaces and Newton-Schulz iteration to reduce complexity.

\section*{Acknowledgements}
This work is supported by National Key Research and Development Program of China (under Grant No. 2023YFB2903904) and Beijing Science and Technology Program (No.Z251100008125003).

% \newpage
\bibliography{main}
\bibliographystyle{tmlr}

\newpage
\appendix

\section{Related work}\label{sec:related-work}

\subsection{GNN and its Training Instability}

Graph neural networks (GNNs)~\citep{kipf2016semi,gilmer2017neural} learn node representations through message passing, but deep propagation is often limited by oversquashing and oversmoothing. Oversquashing was highlighted by \citet{alon2020bottleneck}, and \citet{topping2021understanding} connected it to Jacobian-based sensitivity and proposed curvature-inspired rewiring. Subsequent work has alleviated oversquashing by improving spectral gap or connectivity properties, including expander-style rewiring~\citep{banerjee2022oversquashing}, FoSR~\citep{Karhadkar2023}, DiffWire~\citep{arnaiz2022diffwire}, total-resistance-based rewiring~\citep{black2023understanding}, commute-time analysis~\citep{di2023over}, locality-preserving rewiring~\citep{barbero2023locality}, and multi-hop architectures such as Drew~\citep{gutteridge2023drew}.

Oversmoothing refers to the tendency of node representations to become indistinguishable as depth increases~\citep{li2018deeper, oono2019graph}. This effect has been studied empirically and theoretically~\citep{kipf2016semi,wu2020comprehensive}, with remedies including structure modification such as DropEdge and diffusion-based sparsification~\citep{rong2019dropedge,gasteiger2019diffusion}, normalization methods such as PairNorm~\citep{zhao2020pairnorm}, and architectural changes such as residual connections, Jumping Knowledge, and hierarchical pooling~\citep{li2019deepgcns, chen2020simple, xu2018representation, gao2019graph}. Recent work also studies oversquashing and oversmoothing jointly~\citep{jamadandi2024spectral,arroyo2025vanishing}.

\subsection{Line graphs and edge-centric GNNs.}
Graph neural networks on line graphs and related edge-centric formulations have been studied to better align representation learning with edge-level tasks such as link prediction. In a line graph, each edge in the original graph is represented as a node, allowing message passing to operate directly over links rather than first learning endpoint embeddings and then applying a separate decoder. Early dual or edge-node co-embedding architectures, such as Dual-Primal Graph Convolutional Networks~\citep{monti2018dual} and CensNet~\citep{jiang2019censnet}, propagate information over both vertex- and edge-level structures, with CensNet explicitly using the line graph to switch the roles of nodes and edges. Edge-aware architectures such as NENN~\citep{yang2020nenn} and EGAT~\citep{wang2021egat} further model interactions between node and edge features, although they are not primarily designed as line-graph link-prediction methods. For link prediction, SEAL~\citep{zhang2018link} established the effectiveness of extracting enclosing subgraphs around candidate links and learning structural heuristics with a GNN, but it treats each candidate as a graph-classification instance. LGLP~\citep{cai2022line} instead directly transforms the enclosing subgraph of a candidate link into a line graph, reducing link prediction to node classification on the corresponding edge-node. LGCL~\citep{zhang2023line} extends this direction by introducing contrastive learning between the original subgraph view and the line-graph view, demonstrating that line-graph representations capture complementary edge-neighborhood information useful for predicting missing links. Recent preprint work has also explored directed link prediction by constructing directed line graphs and fusing local and global features~\citep{zhang2025directed}, suggesting a possible extension of line-graph GNNs beyond undirected settings. In parallel, strong link-prediction models such as WalkPooling~\citep{pan2022neural} and ELPH/BUDDY~\citep{chamberlain2023graph} learn subgraph-aware or edge-centric representations without explicitly materializing the full line graph, offering more scalable alternatives when line-graph construction becomes expensive around high-degree nodes. Overall, existing work suggests that line-graph GNNs are attractive for link prediction because they make edges the primary objects of message passing, but their practical use must balance task alignment against scalability, directionality, multigraph handling, and the cost of constructing dense edge-to-edge neighborhoods.

% Prior work studies graph instability mainly through rewiring, diffusion, multi-hop propagation, or parameter-level orthogonality. Our focus is complementary: we target the propagation operator itself. To avoid repeating background material, the next section uses only the minimal preliminaries needed for the subsequent construction and analysis.

\subsection{Unitarity in Deep Learning}\label{ssec:unn}

Unitarity has been widely used to stabilize deep models. In recurrent networks, unitary and orthogonal parameterizations such as uRNN, EUNN, scoRNN, expRNN, projUNN, and LRU improve long-range signal propagation by controlling the spectrum of the transition operator~\citep{arjovsky2016unitary,jing2017tunable,helfrich2018orthogonal,lezcano2019cheap,kiani2022projunn,orvieto2023resurrecting}. Related recurrent architectures, including RWKV and Mamba, share the broader goal of maintaining stable long-range dynamics~\citep{peng2023rwkv,gu2023mamba}.

Beyond RNNs, orthogonality or unitarity has also been explored in convolutional and graph models. Unitary CNNs constrain convolutional filters through structured parameterizations or Cayley/Lie-algebra methods~\citep{sedghi2018singular,li2019preventing,singla2021skew,trockman2021orthogonalizing}, while Ortho-GConv imposes orthogonality on the feature transformation matrix in GNNs~\citep{guo2022orthogonal}. Closely related to our work, \citet{kiani2024unitary} construct unitary graph and group convolutions through matrix exponentials such as $\exp(i\mathbf{A})$, yielding norm-preserving propagation operators with strong empirical performance. GUMP is complementary in both construction and scope: rather than applying a matrix-exponential unitary convolution on the original graph, we transform the input graph into an Eulerian line-graph construction and compute an input-dependent unitary propagation matrix by polar projection while preserving the admissible sparse edge-to-edge transition structure. Orthogonal initialization and optimizers such as Muon further illustrate the usefulness of unitarity at the parameter level~\citep{saxe2013exact,balduzzi2017shattered,jordan2024muon}. More broadly, our setting differs from parameter-level unitarity because the propagation operator is graph-dependent rather than a learned parameter, so enforcing unitarity requires graph transformation and projection rather than only parameter-level constraints.

%\footnote{$\surd$ +qm+ move to Section~\ref{ssec:compare} instead.}
% Unlike Ortho-GConv, GUMP addresses the issue of ill-posed gradient caused by oversquashing by imposing restrictions on the unitarity of the adjacency matrix,
% which is not taken into account in previous works.
% Moreover, enforcing unitarity on adjacency matrix is more challenging than previous works on UNN,
% since the adjacency matrix depends on the input graph and is not a parameter of GNN.

\section{Preliminaries}\label{app:pre}

% \subsection{Graph theory}
\begin{definition}[Line graph]\label{def:line-graph}
    Given a graph $G$, its line graph $\mathsf{L}(G)$ is a graph such that
    \begin{itemize}[leftmargin=*]
        \item each vertex of $\mathsf{L}(G)$ represents an edge of $G$, i.e., $\mathsf{V}[\mathsf{L}(G)] = \mathsf{E}[G]$;
        \item two vertices of $\mathsf{L}(G)$ are adjacent if and only if their corresponding edges share a common endpoint in $G$, i.e., $\mathsf{E}[\mathsf{L}(G)] = \{((i,j), (j,k))\in \mathsf{V}[\mathsf{L}(G)] \times \mathsf{V}[\mathsf{L}(G)] \mid (i,j), (j,k)\in \mathsf{E}[G]\}$.
    \end{itemize}
\end{definition}

\begin{definition}[Eulerian graph]\label{def:eulerian-graph}
    An Eulerian graph $G$ is a graph containing an Eulerian cycle, i.e., there is a trail in $G$ that starts and ends on the same vertex and visits every edge exactly once.
\end{definition}

\begin{definition}[Permutation matrix]
    A permutation matrix $\mathbf P\in \mathbb{R}^{n\times n}$ is a square binary matrix that has exactly one entry of 1 in each row and each column with all other entries 0.
\end{definition}

\begin{theorem}\label{thm:permutation-matrix}
    Every permutation matrix is orthogonal, i.e., if $\mathbf P$ is a permutation matrix, $\mathbf P^\top \mathbf P = \mathbf P \mathbf P^\top = \mathbf I$.
\end{theorem}

\section{Proof}\label{app:proof}

% {\color{purple}
% \subsection{Theory of vanilla message passing}\label{app:formal-theory}

% We provide a formal theory of \cref{thm:informal} to show that GUMP can alleviate oversquashing in \cref{thm:unitary-row-energy}.
% Our theory is based on the GNN model from \cref{fig:gump}(a) with the activation function being ReLU.
% GUMP is analyzed with $\mathbf{A}$ being unitary and vanilla message passing is analyzed with $\mathbf{A}$ being the normalized adjacency matrix.
% Motivated by \citet{xu2018representation}, we analyze with the expected Jacobian.
% We theoretically show the Jacobian of oversquashing in GUMP do not exponentially decay with respect to the number of GNN layers. 

% \begin{theorem}[GUMP]
% The expected Jacobian for GUMP, i.e., $\mathbf{A}$ is unitary in GNN, is approximately in the order of 
% $\mathbb{E}\big[ \partial{\mathbf{h}_i^{(L)}} / \partial{\mathbf{x}_s} \big] = \mathcal{O}( 1 )$.
% \end{theorem}

% Since oversquashing is mainly caused by decay/explosion of Jacobian, \cref{thm:unitary-row-energy} indicates that the Jacobian of GUMP does not change exponentially with respect to $L$ and thus alleviates oversquashing.
%\footnote{$\surd$ +qm+ describe ur theorm first.
%	write sth here.
%	then compare with previous work below.
%	Place them together make these two theorms are of the same importance to ur work.
%	Or,
%	if theorem 3.6 can be derived from 3.5,
%	you can change it to corollary.}

\subsection{Proof of Proposition~1}
Proposition~1 is proved based on the following theorem and lemma.
Theorem~\ref{thm:existence} is a direct result from Theorem 3 in \citet{severini2003digraph}. Lemma~\ref{lem:eulerian-graph} is a well-known result in graph theory, which can be found in Theorem 1.7.2 of \citet{bang2008digraphs}.
\begin{theorem}[Existence of unitary adjacency matrix]\label{thm:existence}
    Let $G$ be a single-connected digraph. Its line graph $\mathsf{L}(G)$ (Definition~\ref{def:line-graph}) is the digraph of a unitary matrix if and only if $G$ is Eulerian (Definition~\ref{def:eulerian-graph}).
\end{theorem}

\begin{lemma}[A special Eulerian graph]\label{lem:eulerian-graph}
    A digraph graph is Eulerian if and only if it is connected and the in-degree and out-degree are equal at each vertex.
\end{lemma}

\begin{proof}[Proof of Proposition~1]
    In Algorithm~1, the undirected edges in $G$ are split into two directed edges in $G'$.
    Therefore, the in-degree and out-degree of each vertex in $G'$ are equal.
    {We first consider the case where $G$ is connected. Replacing each undirected edge by the pair of directed edges $(i,j)$ and $(j,i)$ preserves connectivity of the underlying graph, so $G'$ is connected as well. By Lemma~\ref{lem:eulerian-graph}, $G'$ is Eulerian, and Theorem~\ref{thm:existence} yields a unitary adjacency matrix $\mathbf{U}$ such that $\mathsf{S}[\mathbf{U}] = \mathbf{A}[\mathsf{L}(G')]$.}

    {Now suppose that $G$ is disconnected. Let $G_1,\ldots,G_m$ be the connected components of $G$ that contain at least one edge, and let $G_1',\ldots,G_m'$ be the corresponding connected components of $G'$ after the edge-splitting step. The same argument as above shows that each $G_r'$ is connected and satisfies $d_{\mathrm{in}}(v)=d_{\mathrm{out}}(v)$ at every vertex, so Lemma~\ref{lem:eulerian-graph} implies that each $G_r'$ is Eulerian. Theorem~\ref{thm:existence} then gives a unitary adjacency matrix $\mathbf{U}_r$ for each $\mathsf{L}(G_r')$. Since the line graph of a disjoint union is the disjoint union of the corresponding line graphs, $\mathsf{L}(G')$ is block diagonal up to vertex ordering with blocks $\mathsf{L}(G_1'),\ldots,\mathsf{L}(G_m')$; isolated vertices of $G$ contribute no vertices to $\mathsf{L}(G')$, but their node features remain available through GUMP's skip/concatenation path. Hence $\mathbf{U}:=\mathsf{diag}(\mathbf{U}_1,\ldots,\mathbf{U}_m)$ is unitary and satisfies $\mathsf{S}[\mathbf{U}] = \mathbf{A}[\mathsf{L}(G')]$.}
\end{proof}

\subsection{Proof of Proposition~\ref{prop:permutation-equivariance}}
\begin{lemma}\label{lem:perumte-unitary}
    For any unitary matrix $\mathbf{U}$, given two permutation matrices $\mathbf{P}_1$ and $\mathbf P_2$, $\mathbf{P}_1 \mathbf{U} \mathbf{P}_2^\top$ is also unitary.
\end{lemma}
\begin{proof}
    Let $\hat{\mathbf{U}}= \mathbf{P}_1 \mathbf{U}\mathbf{P}_2^\top$. Then, we have
    \begin{align*}
        & \hat{\mathbf U} \hat{\mathbf U}^\dagger = \mathbf{P}_1 \mathbf{U}\mathbf{P}_2^\top \mathbf{P}_2 \mathbf{U}^\dagger \mathbf{P}_1^\top = \mathbf{P}_1 \mathbf{U} \mathbf{U}^\dagger \mathbf{P}_1^\top = \mathbf{P}_1 \mathbf{P}_1^\top = \mathbf{I}, \\
        & \hat{\mathbf U}^\dagger \hat{\mathbf U} = \mathbf{P}_2 \mathbf{U}^\dagger \mathbf{P}_1^\top \mathbf{P}_1 \mathbf{U} \mathbf{P}_2^\top = \mathbf{P}_2 \mathbf{U}^\dagger \mathbf{U} \mathbf{P}_2^\top = \mathbf{P}_2 \mathbf{P}_2^\top = \mathbf{I}
    \end{align*}
    which proves $\mathbf{P}_1 \mathbf{U}\mathbf{P}_2^\top$ is unitary.
\end{proof}

{
\begin{proof}[Proof of Proposition~\ref{prop:permutation-equivariance}]
    Since $\mathbf{A}_w$ is full-rank, the polar factor is given by
    $\mathsf{U}[\mathbf{A}_w]=\mathbf{A}_w(\mathbf{A}_w^\dagger \mathbf{A}_w)^{-1/2}$.
    The full-rank assumption is used here to ensure that $\mathbf{A}_w^\dagger \mathbf{A}_w$ is positive definite, so that $(\mathbf{A}_w^\dagger \mathbf{A}_w)^{-1/2}$ exists and the polar factor $\mathbf{A}_w(\mathbf{A}_w^\dagger \mathbf{A}_w)^{-1/2}$ is uniquely defined. Row and column permutations preserve rank, so the same condition holds for $\mathbf{P}_1\mathbf{A}_w\mathbf{P}_2^\top$.
    Let $\hat{\mathbf A}_w=\mathbf{P}_1 \mathbf{A}_w \mathbf{P}_2^\top$. Then
    \begin{align*}
        \hat{\mathbf A}_w^\dagger \hat{\mathbf A}_w
        &= \mathbf{P}_2 \mathbf{A}_w^\dagger \mathbf{P}_1^\top \mathbf{P}_1 \mathbf{A}_w \mathbf{P}_2^\top \\
        &= \mathbf{P}_2 (\mathbf{A}_w^\dagger \mathbf{A}_w) \mathbf{P}_2^\top.
    \end{align*}
    Because permutation similarity preserves matrix functions,
    \begin{equation*}
        (\hat{\mathbf A}_w^\dagger \hat{\mathbf A}_w)^{-1/2}
        = \mathbf{P}_2 (\mathbf{A}_w^\dagger \mathbf{A}_w)^{-1/2} \mathbf{P}_2^\top.
    \end{equation*}
    Therefore,
    \begin{align*}
        \mathsf{U}[\hat{\mathbf A}_w]
        &= \hat{\mathbf A}_w (\hat{\mathbf A}_w^\dagger \hat{\mathbf A}_w)^{-1/2} \\
        &= \mathbf{P}_1 \mathbf{A}_w \mathbf{P}_2^\top
        \mathbf{P}_2 (\mathbf{A}_w^\dagger \mathbf{A}_w)^{-1/2} \mathbf{P}_2^\top \\
        &= \mathbf{P}_1 \mathbf{A}_w (\mathbf{A}_w^\dagger \mathbf{A}_w)^{-1/2} \mathbf{P}_2^\top \\
        &= \mathbf{P}_1 \mathsf{U}[\mathbf{A}_w] \mathbf{P}_2^\top,
    \end{align*}
    which proves the proposition.
\end{proof}
}

\subsection{Proof of Proposition~\ref{prop:block-support}}
\begin{proof}
    Let $G^\prime={(V,E^\prime)}$. Index the rows and columns of $\mathbf{A}_w$ and $\mathbf{A}_{\mathsf L}$ by directed edges of {$E^\prime$}. By Definition~\ref{def:line-graph}, for two directed edges {$e=(u,v)$ and $f=(x,y)$}, the entry $(\mathbf{A}_{\mathsf L})_{ef}$ is one exactly when {$v=x$}. Thus $(\mathbf{A}_w)_{ef}$ can be nonzero only when $e\in E_{\mathrm{in}}(v)$ and $f\in E_{\mathrm{out}}(v)$ for the same vertex $v\in {V}$.

    Let $\mathbf{P}_{\mathrm{in}}$ be the permutation that groups rows by the sets $E_{\mathrm{in}}(v)$ and let $\mathbf{P}_{\mathrm{out}}$ be the permutation that groups columns by the sets $E_{\mathrm{out}}(v)$. Under these permutations, rows associated with $E_{\mathrm{in}}(v)$ interact only with columns associated with $E_{\mathrm{out}}(v)$, so
    \[
    \mathbf{D}:=\mathbf{P}_{\mathrm{in}}\mathbf{A}_w\mathbf{P}_{\mathrm{out}}^\top
    = \mathsf{diag}\!\big(\mathbf{B}_v\big)_{v\in {V}}.
    \]
    Each block $\mathbf{B}_v$ contains all transitions $(u,v)\to (v,w)$ in the line graph. The corresponding block of the binary adjacency $\mathbf{A}_{\mathsf L}$ is the all-ones matrix $\mathbf{J}_v=\mathbf{1}_{d(v)\times d(v)}$, because every incoming directed edge to $v$ can be followed by every outgoing directed edge from $v$ in the line graph. Because Algorithm~\ref{alg:transformation} inserts both orientations of every undirected edge, $|E_{\mathrm{in}}(v)|=|E_{\mathrm{out}}(v)|$, hence $\mathbf{B}_v$ is square.

    Since $\mathbf{D}$ is block diagonal,
    \[
    \mathbf{D}^\dagger \mathbf{D}
    = \mathsf{diag}\!\big(\mathbf{B}_v^\dagger \mathbf{B}_v\big)_{v\in {V}}.
    \]
    Matrix functions preserve block diagonality, so under the full-rank assumption,
    \[
    (\mathbf{D}^\dagger \mathbf{D})^{-1/2}
    = \mathsf{diag}\!\big((\mathbf{B}_v^\dagger \mathbf{B}_v)^{-1/2}\big)_{v\in {V}}.
    \]
    Therefore,
    \begin{align*}
        \mathsf{U}[\mathbf{D}]
        &= \mathbf{D}(\mathbf{D}^\dagger \mathbf{D})^{-1/2} \\
        &= \mathsf{diag}\!\big(\mathbf{B}_v(\mathbf{B}_v^\dagger \mathbf{B}_v)^{-1/2}\big)_{v\in {V}} \\
        &= \mathsf{diag}\!\big(\mathsf{U}[\mathbf{B}_v]\big)_{v\in {V}}.
    \end{align*}
    By Proposition~\ref{prop:permutation-equivariance},
    \[
    \mathsf{U}[\mathbf{D}]
    = \mathsf{U}[\mathbf{P}_{\mathrm{in}}\mathbf{A}_w\mathbf{P}_{\mathrm{out}}^\top]
    = \mathbf{P}_{\mathrm{in}}\mathsf{U}[\mathbf{A}_w]\mathbf{P}_{\mathrm{out}}^\top.
    \]
    {Hence $\mathsf{U}[\mathbf{A}_w] = \mathbf{P}_{\mathrm{in}}^\top \mathsf{U}[\mathbf{D}] \mathbf{P}_{\mathrm{out}}$ has the same row/column block structure as $\mathbf{A}_w$. Since each projected block $\mathsf{U}[\mathbf{B}_v]$ is contained in the corresponding all-ones support block $\mathbf{J}_v$, every nonzero entry of $\mathsf{U}[\mathbf{A}_w]$ lies on an admissible transition of $\mathsf{L}(G^\prime)$. Therefore,}
    \[
    {\mathsf{S}[\mathsf{U}[\mathbf{A}_w]] \subseteq \mathsf{S}[\mathbf{A}_{\mathsf L}].}
    \]
    {If each block polar factor $\mathsf{U}[\mathbf{B}_v]$ is fully supported, this inclusion becomes equality. This condition is generic for full-rank blocks whose entries are produced by continuous learned weights: each entry of $\mathsf{U}[\mathbf{B}_v]$ is a real-analytic function of the entries of $\mathbf{B}_v$ on the full-rank domain, and it is not identically zero because one can choose a fully supported unitary block as $\mathbf{B}_v$. Hence the weights that make a specific projected entry exactly zero form a measure-zero set.}

    For the Newton-Schulz implementation, {let $\mathbf{X}_{v,t}$ denote the $t$-th iterate for block $\mathbf{B}_v$,} initialized as $\mathbf{X}_{v,0}=\mathbf{B}_v/\|\mathbf{B}_v\|_F$. If $\mathbf{X}_{v,t}$ is blockwise supported, then $\mathbf{X}_{v,t}^\top \mathbf{X}_{v,t}$ and the polynomial update in Algorithm~\ref{alg:efficient-GUMP} remain within the same block. By induction, every iterate is block diagonal under the same permutations, so no off-support fill-in is introduced during the iteration either.
\end{proof}

\subsection{Proofs for Section~\ref{sec:gump}}
Our proof is based on the GNN model from Figure~2(a) with the activation function being ReLU.
GUMP is analyzed with $\mathbf{A}$ being unitary and vanilla message passing is analyzed with $\mathbf{A}$ being the normalized adjacency matrix.

{
\paragraph{Proof of Proposition~\ref{prop:expected-jacobian-factorization}.}
\begin{proof}
    Denote by $\mathbf{f}_i^{(l)}$ the pre-activated feature of $\mathbf{h}_i^{(l)}$, i.e., $\mathbf{f}_i^{(l)}=\sum\nolimits_{z \in \mathcal{N}(i)} \mathbf{A}_{iz} \mathbf{h}_z^{(l-1)}\mathbf{W}_l$, for any $l=1\cdots L$, we have
    \begin{equation*}
        \frac{\partial \mathbf{h}_i^{(l)}}{\partial \mathbf{h}_s^{(0)}} = \mathsf{diag}\left(1_{\mathbf{f}_i^{(l)} > 0}\right) \cdot \left( \sum_{z\in \mathcal{N}(i)} \mathbf{A}_{iz} \frac{\partial \mathbf{h}_z^{(l-1)}}{\partial \mathbf{h}_s^{(0)}} \right) \cdot \mathbf{W}_l^\top.
    \end{equation*}
    By the chain rule, we get 
    \begin{align*}
        \frac{\partial \mathbf{h}_i^{(L)}}{\partial \mathbf{h}_s^{(0)}} = & \sum_{p=1}^{\Psi} \left[ \frac{\partial \mathbf{h}_i^{(L)}}{\partial \mathbf{h}_s^{(0)}} \right]_p \\
        = & \sum_{p=1}^{\Psi} \prod_{l=L}^{1} \mathsf{diag}\left(1_{\mathbf{f}_{v_p^l}^{(l)} > 0}\right) \mathbf{A}_{v_p^l v_p^{l-1}} \mathbf{W}_l^\top.  \\
    \end{align*}
    Here, $\Psi$ is the total number of paths $v_p^L v_p^{L-1}\cdots v_p^1 v_p^0$ of length $L+1$ from $v_p^0=s$ to $v_p^L=i$. For $l=1\cdots L-1$, $v_p^{l-1}\in \mathcal{N}(v_p^l)$.

    For each path $p$, the derivative $[\partial \mathbf{h}_i^{(L)} / \partial \mathbf{h}_s^{(0)} ]_p$ represents a directed acyclic computation graph. At a layer $l$, we can express an entry of the derivative as
    \begin{equation*}
        \left[ \frac{\partial \mathbf{h}_i^{(L)}}{\partial \mathbf{h}_s^{(0)}} \right]_p^{(m,n)} = \prod_{l=L}^{1} \mathbf{A}_{v_p^l v_p^{l-1}} \sum_{q=1}^{\Phi} Z_q \prod_{l=L}^1  {w}_q^{(l)},
    \end{equation*} 
    where $\Phi$ is the number of paths $q$ from the input neurons to the output neuron $(m,n)$, in the computation graph of $[\partial \mathbf{h}_i^{(L)} / \partial \mathbf{h}_s^{(0)} ]_p$.
    For each layer $l$, ${w}_q^{(l)}$ is the entry of $\mathbf{W}_l^\top$ that is used in the $q$-th path.
    Finally, $Z_q\in\{0,1\}$ represents whether the $q$-th path is active ($Z_q=1$) or not ($Z_q=0$) as a result of ReLU activation of the entries of $\mathbf{f}_{v_p^l}^{(l)}$'s on the $q$-th path.

    Under the assumption that $Z_q$ contributes the path-level gate factor $\rho_L$.
    Because of $\mathbb{P}[Z_q=1]=\rho_L, \forall q$, we have 
    \begin{equation*}
        \mathbb{E}\left[ \left[ \frac{\partial \mathbf{h}_i^{(L)}}{\partial \mathbf{h}_s^{(0)}} \right]_p^{(m,n)} \right] = \rho_L \prod_{l=L}^{1} \mathbf{A}_{v_p^l v_p^{l-1}} \sum_{q=1}^{\Phi} \prod_{l=L}^{1} {w}_q^{(l)}.
    \end{equation*}
    Then, the expected Jacobian is
\begin{equation*}
        \mathbb{E}\left[ \frac{\partial{\mathbf{h}_i^{(L)}}}{\partial{\mathbf{x}_s}} \right] =  \sum_{p=1}^{\Psi}\mathbb{E}\left[ \left[ \frac{\partial \mathbf{h}_i^{(L)}}{\partial \mathbf{h}_s^{(0)}} \right]_p \right] = \rho_L \prod_{l=L}^1 \mathbf{W}_l^\top \left(\mathbf{A}^{L}\right)_{is}.
    \end{equation*}
\end{proof}
}

{
\paragraph{Average-over-sources identity.}
For the following two theorems, let $N$ denote the size of the propagation matrix, fix a target node $i\in[N]$, and sample the source node $s\sim \mathrm{Unif}([N])$.
This metric is natural for our analysis because Proposition~\ref{prop:expected-jacobian-factorization} factorizes the expected Jacobian into the cumulative weight term and the scalar graph-propagation term $(\mathbf{A}^{L})_{is}$. Averaging over $s$ therefore measures the influence of a typical source on the fixed target node $i$, while the squared magnitude removes sign or phase cancellations that can make individual entries uninformative. The resulting quantity is exactly the row energy of the propagation matrix at node $i$.
With the notation introduced in the main text, this quantity is $I_i(L;\mathbf{P})$.
\begin{lemma}\label{lem:avg_sources_identity}
    For any $\mathbf{P}\in\mathbb{C}^{N\times N}$ and any $L\ge 0$,
    \begin{equation*}
        \mathbb{E}_{s}\!\left[\left|(\mathbf{P}^{L})_{is}\right|^{2}\right]
        = \frac{1}{N}\sum_{s=1}^{N}\left|(\mathbf{P}^{L})_{is}\right|^{2}
        = \frac{1}{N}\left\|\mathbf{e}_{i}^{\top}\mathbf{P}^{L}\right\|_{2}^{2}
        = \frac{1}{N}\left(\mathbf{P}^{L}(\mathbf{P}^{L})^{\dagger}\right)_{ii}.
    \end{equation*}
\end{lemma}
\begin{proof}
    By definition,
    \begin{equation*}
        \mathbb{E}_{s}\!\left[\left|(\mathbf{P}^{L})_{is}\right|^{2}\right]
        = \frac{1}{N}\sum_{s=1}^{N}\left|\mathbf{e}_{i}^{\top}\mathbf{P}^{L}\mathbf{e}_{s}\right|^{2}
        = \frac{1}{N}\left\|\mathbf{e}_{i}^{\top}\mathbf{P}^{L}\right\|_{2}^{2}.
    \end{equation*}
    Moreover,
    \begin{equation*}
        \left\|\mathbf{e}_{i}^{\top}\mathbf{P}^{L}\right\|_{2}^{2}
        = \mathbf{e}_{i}^{\top}\mathbf{P}^{L}(\mathbf{P}^{L})^{\dagger}\mathbf{e}_{i}
        = \left(\mathbf{P}^{L}(\mathbf{P}^{L})^{\dagger}\right)_{ii},
    \end{equation*}
    which proves the claim.
\end{proof}
}

{
\paragraph{Proof of Theorem~\ref{thm:unitary-row-energy}.}
\begin{proof}
    Since $\mathbf{A}$ is unitary, $\mathbf{A}^{L}$ is unitary for every $L$, and hence
    \begin{equation*}
        \mathbf{A}^{L}(\mathbf{A}^{L})^{\dagger} = \mathbf{I}.
    \end{equation*}
    Applying Lemma~\ref{lem:avg_sources_identity} with $\mathbf{P}=\mathbf{A}$ gives
    \begin{equation*}
        \mathbb{E}_{s}\!\left[\left|(\mathbf{A}^{L})_{is}\right|^{2}\right]
	        = \frac{1}{N}\left(\mathbf{A}^{L}(\mathbf{A}^{L})^{\dagger}\right)_{ii}
	        = \frac{1}{N}.
    \end{equation*}
    Since the left-hand side is $I_i(L;\mathbf{A})$, this proves $I_i(L;\mathbf{A})=1/N$.
\end{proof}
}

\subsection{Proof of Theorem~\ref{thm:vanilla-spectral-decay}}
\label{app:vanilla-spectral-decay-proof}

{
\begin{proof}
    Because $\hat{\mathbf A}$ is real symmetric, it admits the orthonormal eigendecomposition
    \begin{equation*}
        \hat{\mathbf A}
        = \sum_{j=1}^{N} \lambda_j v_j v_j^\top,
        \qquad
        \hat{\mathbf A}^{L}
        = \sum_{j=1}^{N} \lambda_j^{L} v_j v_j^\top.
    \end{equation*}
    Since $\mathbf{\Pi}=v_1v_1^\top$, we also have
    \begin{equation*}
        \hat{\mathbf A}^{L}-\mathbf{\Pi}
        = \sum_{j=2}^{N} \lambda_j^{L} v_j v_j^\top.
    \end{equation*}

    Applying Lemma~\ref{lem:avg_sources_identity} with $\mathbf{P}=\hat{\mathbf A}$ and using symmetry gives
    \begin{equation*}
        \mathbb{E}_{s}\!\left[\left|(\hat{\mathbf A}^{L})_{is}\right|^{2}\right]
        = \frac{1}{N}\left(\hat{\mathbf A}^{L}(\hat{\mathbf A}^{L})^\top\right)_{ii}
        = \frac{1}{N}\left(\hat{\mathbf A}^{2L}\right)_{ii}.
    \end{equation*}
    Expanding $\hat{\mathbf A}^{2L}$ in the eigenbasis yields
    \begin{equation*}
        \left(\hat{\mathbf A}^{2L}\right)_{ii}
        = \sum_{j=1}^{N} \lambda_j^{2L} v_{j,i}^{2}
        = v_{1,i}^{2} + \sum_{j=2}^{N} \lambda_j^{2L} v_{j,i}^{2},
    \end{equation*}
    which proves the exact decomposition. Since $|\lambda_j|\le c$ for $j\ge 2$ and $\sum_{j=2}^{N} v_{j,i}^{2}\le 1$,
    \begin{equation*}
        0 \le \frac{1}{N}\sum_{j=2}^{N} v_{j,i}^{2}\lambda_j^{2L}
        \le \frac{1}{N}c^{2L}\sum_{j=2}^{N} v_{j,i}^{2}
        \le \frac{1}{N}c^{2L}.
    \end{equation*}

    For the centered quantity, apply Lemma~\ref{lem:avg_sources_identity} to $\mathbf{P}=\hat{\mathbf A}^{L}-\mathbf{\Pi}$:
    \begin{equation*}
        \mathbb{E}_{s}\!\left[\left|\delta_L(i,s)\right|^{2}\right]
        = \frac{1}{N}\left((\hat{\mathbf A}^{L}-\mathbf{\Pi})(\hat{\mathbf A}^{L}-\mathbf{\Pi})^\top\right)_{ii}.
    \end{equation*}
    Using the eigendecomposition above and orthonormality of $\{v_j\}$, we obtain
    \begin{equation*}
        \mathbb{E}_{s}\!\left[\left|\delta_L(i,s)\right|^{2}\right]
        = \frac{1}{N}\sum_{j=2}^{N} v_{j,i}^{2}\lambda_j^{2L}
        \le \frac{1}{N}c^{2L}.
    \end{equation*}
    For the lower bound, keep only the terms with $|\lambda_j|=c$:
    \begin{equation*}
        \mathbb{E}_{s}\!\left[\left|\delta_L(i,s)\right|^{2}\right]
        \ge \frac{1}{N}\sum_{j:\,|\lambda_j|=c} v_{j,i}^{2} c^{2L}
        = \frac{1}{N}\alpha_i c^{2L}.
    \end{equation*}
    Hence the non-stationary source-averaged influence decays at rate $\Theta(c^{2L}/N)$ whenever $\alpha_i>0$.
\end{proof}
}

% \section{Extension of GUMP}

\clearpage

\section{More Experimental Results}\label{app:exp}

\paragraph{Statistics of datasets} The statistics of datasets used in experiments are shown in Table~\ref{tab:dataset-statistics}.
\begin{table}[H]
    \centering
    % \scriptsize
    \caption{Statistics of datasets.}\label{tab:dataset-statistics}
    \begin{tabular}{c|cccc}
        \toprule
                    & \#graphs & Avg. nodes & Avg. edges & Task type            \\
                    \midrule
    Mutag           & 188      & 17.9    & 39.6    & Graph Classification \\
    Proteins        & 1,113    & 39.1    & 145.6   & Graph Classification \\
    Enzymes         & 600      & 32.6    & 124.3   & Graph Classification \\
    NCI1            & 4110     & 29.87   & 32.30   & Graph Classification \\
    NCI109          & 4127     & 29.68   & 32.13   & Graph Classification \\
    Peptides-func   & 15,535   & 150.94  & 307.30  & Graph Classification \\
    Peptides-struct & 15,535   & 150.94  & 307.30  & Graph Regression     \\ \bottomrule  
    \end{tabular}
\end{table}

\paragraph{Hyperparameters of GUMP} The hyperparameters of GUMP for both synthetic and real datasets are shown in Table~\ref{tab:hyperparameters}.

\begin{table}[H]
    \centering
    \caption{Hyperparameters of GUMP for datasets in experiments. $\text{layer}_{\text{GUMP}}$, $\text{lr}_{\text{base}}$, $\text{wd}_{\text{base}}$, $\text{lr}_{\text{GUMP}}$, $\text{wd}_{\text{GUMP}}$, drop., $d'$, $d$, batch size, $\text{layer}_{\text{base}}$, opt., sched., and epoch denotes the number of layers of GUMP, the learning rate of base GNN, weight decay of base GNN, the learning rate of GUMP, weight decay of GUMP, dropout rate, dimension of calculating (1), hidden dimension of GNN, batch size, number of layers of base GNN, optimizer, scheduler, and number of epochs, respectively.}\label{tab:hyperparameters}
    % \tiny
    % \renewcommand{\arraystretch}{1.0}
    \footnotesize 
    \setlength\tabcolsep{2pt}
    \begin{tabular}{c|ccccccccccccc}
        \toprule
     & $\text{layer}_{\text{GUMP}}$ & $\text{lr}_{\text{base}}$ & $\text{wd}_{\text{base}}$ & $\text{lr}_{\text{GUMP}}$ & $\text{wd}_{\text{GUMP}}$ & drop. & $d'$ & $d$ & batch size & $\text{layer}_{\text{base}}$ & opt. & sched. & epoch \\ \midrule
     CrossedRing           & - & $10^{-4}$ & $10^{-6}$ & $10^{-4}$ & $0$       & $0$   & $32$ & $32$  & 20    & 0 & adam & none               & 200 \\
     Ring           & - & $10^{-4}$ & $10^{-6}$ & $10^{-4}$ & $0$       & $0$   & $32$ & $32$  & 20    & 0 & adam & none               & 200 \\
     CliquePath           & - & $10^{-4}$ & $10^{-6}$ & $10^{-4}$ & $0$       & $0$   & $32$ & $32$  & 20    & 0 & adam & none               & 200 \\
     Mutag           & 16 & $10^{-2}$ & $10^{-4}$ & $10^{-4}$ & $0$       & $0$   & $32$ & $64$  & 16    & 5 & adam & none               & 100 \\
    Proteins        & 20 & $10^{-2}$ & $10^{-2}$ & $10^{-4}$ & $10^{-2}$ & $0$   & $32$ & $64$  & 64    & 3 & adam & none               & 100 \\
    Enzymes         & 10 & $10^{-2}$ & $10^{-4}$ & $10^{-4}$ & $0$       & $0$   & $32$ & $64$  & 16    & 1 & adam & none               & 100 \\
    NCI1            & 10  & $10^{-2}$ & $10^{-4}$ & $10^{-4}$ & $0$       & $0$   & $32$ & $64$  & 16    & 1 & adam & none               & 100 \\
    NCI109          & 10  & $10^{-2}$ & $10^{-4}$ & $10^{-4}$ & $0$       & $0$   & $32$ & $64$  & 16    & 1 & adam & none               & 100 \\
    Peptides-func   & 12 & $0.005$   & $0.1$     & $0.1$     & $0.1$     & $0.2$ & $32$ & $256$ & 200   & 3 & adam & cos. & 250 \\
    Peptides-struct & 12 & $0.005$   & $0.1$     & $0.005$   & $0.1$     & $0.2$ & $32$ & $256$ & 200   & 3 & adam & cos. & 250 \\ \bottomrule
    \end{tabular}
\end{table}

{
\subsection{Empirical Block-Rank and Finite-Step Unitarity Diagnostics}
\label{app:unitary-diagnostics}

\paragraph{Setup.}
We add diagnostics for the two numerical conditions most directly related to Proposition~\ref{prop:permutation-equivariance} and Algorithm~\ref{alg:efficient-GUMP}. For each line-graph transition block $\mathbf{B}_v$, we report the full-rank rate under tolerance $10^{-7}$, statistics of the minimum singular value $\sigma_{\min}(\mathbf{B}_v)$, and the finite-step Newton--Schulz residual
\[
  r_{v,K}
  =
  \|\mathbf{X}_{v,K}^{\top}\mathbf{X}_{v,K}-I_{d_v}\|_F/\sqrt{d_v},
\]
where $\mathbf{X}_{v,K}$ is the block returned after $K$ Newton--Schulz iterations. The diagnostic uses the same TU line-graph block partition as the implementation and random continuous 64-dimensional node representations passed through the attention form in \eqref{eq:edge-weight}, rather than raw discrete TU labels. This avoids degeneracies that can arise from discrete labels and probes the generic continuous-representation regime relevant to learned GUMP blocks.

\begin{table}[H]
\centering
\caption{Empirical full-rank and minimum-singular-value statistics for TU transition blocks $\mathbf{B}_v$. Ranks are computed with tolerance $10^{-7}$.}
\label{tab:block-rank-minsv}
\scriptsize
\setlength{\tabcolsep}{4pt}
\begin{tabular}{lrrrrrr}
\toprule
Dataset & \#blocks & max $d_v$ & full-rank (\%) &
min $\sigma_{\min}$ & p1 $\sigma_{\min}$ & med. $\sigma_{\min}$ \\
\midrule
MUTAG    & 3,371   & 4  & 100.00 & $2.50{\times}10^{-6}$ & $1.42{\times}10^{-4}$ & $1.11{\times}10^{-1}$ \\
PROTEINS & 43,466  & 25 & 99.99  & $2.30{\times}10^{-8}$ & $2.07{\times}10^{-5}$ & $2.65{\times}10^{-2}$ \\
ENZYMES  & 19,474  & 9  & 100.00 & $4.45{\times}10^{-7}$ & $1.07{\times}10^{-4}$ & $5.19{\times}10^{-3}$ \\
NCI1     & 122,319 & 4  & 100.00 & $2.07{\times}10^{-7}$ & $8.87{\times}10^{-4}$ & $1.33{\times}10^{-1}$ \\
NCI109   & 122,020 & 5  & 100.00 & $8.16{\times}10^{-7}$ & $6.43{\times}10^{-4}$ & $1.37{\times}10^{-1}$ \\
\bottomrule
\end{tabular}
\end{table}

\begin{table}[H]
\centering
\caption{Finite-step Newton--Schulz residuals $r_{v,K}=\|\mathbf{X}_{v,K}^{\top}\mathbf{X}_{v,K}-I\|_F/\sqrt{d_v}$ for $K=5$ and $K=10$.}
\label{tab:ns-finite-step-residual}
\scriptsize
\setlength{\tabcolsep}{3.2pt}
\begin{tabular}{lrrrrrr}
\toprule
Dataset & med. $r_{v,5}$ & p95 $r_{v,5}$ & max $r_{v,5}$ &
med. $r_{v,10}$ & p95 $r_{v,10}$ & max $r_{v,10}$ \\
\midrule
MUTAG    & $5.98{\times}10^{-2}$ & $7.07{\times}10^{-1}$ & $8.16{\times}10^{-1}$ & $1.09{\times}10^{-7}$ & $5.48{\times}10^{-1}$ & $8.16{\times}10^{-1}$ \\
PROTEINS & $4.98{\times}10^{-1}$ & $8.16{\times}10^{-1}$ & $9.35{\times}10^{-1}$ & $1.89{\times}10^{-7}$ & $5.87{\times}10^{-1}$ & $9.28{\times}10^{-1}$ \\
ENZYMES  & $4.37{\times}10^{-1}$ & $6.20{\times}10^{-1}$ & $8.37{\times}10^{-1}$ & $1.36{\times}10^{-7}$ & $2.61{\times}10^{-1}$ & $7.07{\times}10^{-1}$ \\
NCI1     & $4.86{\times}10^{-3}$ & $6.84{\times}10^{-1}$ & $8.64{\times}10^{-1}$ & $9.78{\times}10^{-8}$ & $9.34{\times}10^{-2}$ & $8.09{\times}10^{-1}$ \\
NCI109   & $5.91{\times}10^{-3}$ & $6.94{\times}10^{-1}$ & $8.65{\times}10^{-1}$ & $9.95{\times}10^{-8}$ & $2.06{\times}10^{-1}$ & $8.14{\times}10^{-1}$ \\
\bottomrule
\end{tabular}
\end{table}

These diagnostics support the generic full-rank condition used by Proposition~\ref{prop:permutation-equivariance}: all five datasets have essentially full-rank transition blocks, with only PROTEINS showing a small number of rank-deficient blocks under the conservative $10^{-7}$ tolerance. For finite-step Newton--Schulz projection, increasing from $K=5$ to $K=10$ reduces the median normalized residual to approximately $10^{-7}$ across all five datasets. The larger p95 and maximum residuals are concentrated in the ill-conditioned tail of blocks, consistent with the lower tail of $\sigma_{\min}(\mathbf{B}_v)$. Since the projection is blockwise, this tail is localized and reflects the usual trade-off between a fixed iteration budget and numerical accuracy; if a stricter tolerance is desired, it can be addressed by increasing $K$ or selectively applying a higher-accuracy block projection to near-singular blocks.
}

\clearpage

\begin{figure}[H]
%    \vspace{-20px}
   \centering
   \subfigure[CrossedRing]{
     \includegraphics[width=0.2\textwidth]{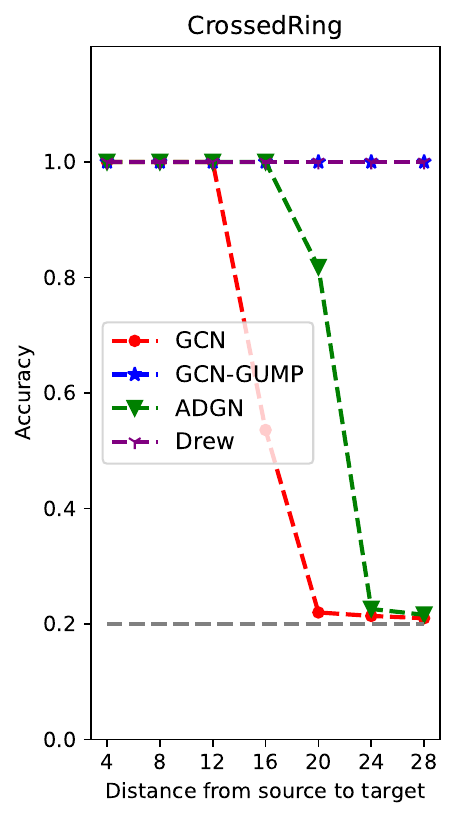}
     \label{fig:app_cross_ring}}
   \hspace{-7px}
   \subfigure[Ring]{
     \includegraphics[width=0.2\textwidth]{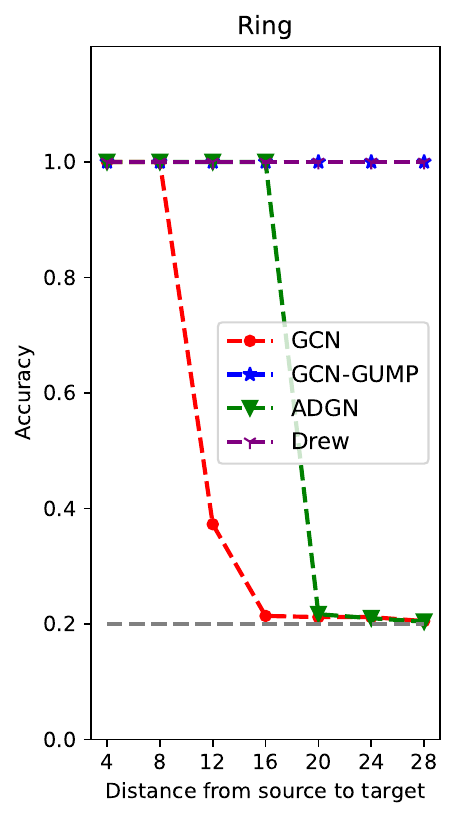}
     \label{fig:app_ring}}
   \hspace{-7px}
   \subfigure[CliquePath]{
     \includegraphics[width=0.2\textwidth]{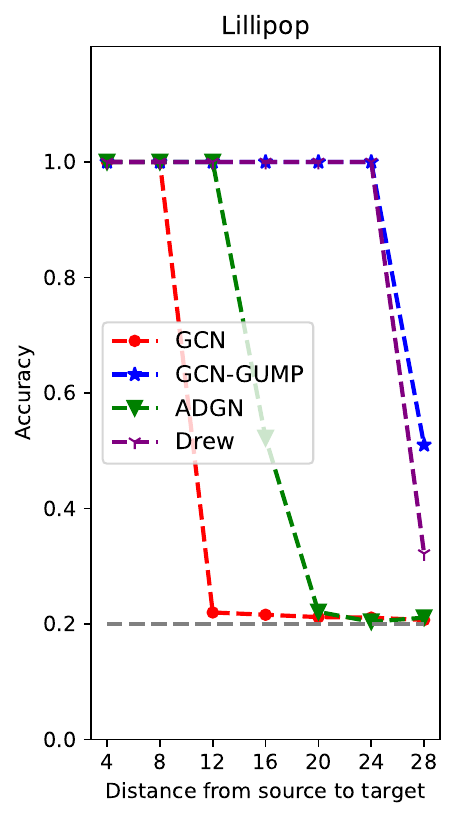}\label{fig:app_lollipop}}
  \vspace{-10px}
   \caption{The performance of GCN and GCN-GUMP on the CrossedRing, Ring, and CliquePath with different distances from source to target.}
   % \vspace{10px}
   \label{fig:app_synthetic}
\end{figure}

\paragraph{Comparison GUMP with more methods} 
For the synthetic datasets, we compare GUMP with Drew and ADGN on synthetic datasets in Figure~\ref{fig:app_synthetic}. The results show that the performances of GUMP and Drew are close, while ADGN performs worse.

We further compare the performances of GUMP by applying different convolution operator. The results are shown in Table~\ref{tab:graph-classification-gcn}. The results show that GUMP can improve the performance of both GCN and GIN, while GIN-GUMP achieves better performance than GCN-GUMP.

\begin{table}[H]
    \centering
    \caption{Graph classification accuracy on the TUDataset. {\bf First}, {\underline{second}}, and {\textit{third}} best results are bold, underlined, and underwaved, respectively.}
    \label{tab:graph-classification-gcn}    % \scriptsize
    % \vspace{-5px}
    \setlength\tabcolsep{2pt}
    % \small
    \begin{tabular}{c|c|ccccc|c}
    \toprule
    Base GNN & Methods & Mutag & Proteins & Enzymes & NCI1 & NCI109 & Rank \\
    \midrule
    \multirow{7}{*}{GCN} & None & 72.15±2.44 & 70.98±0.74 & 27.67±1.16 & 68.74±0.45 & 67.90±0.50 & 4.2 \\
    & None (+layer) & 70.05±1.83 & 69.80±0.99 & 23.63±1.07 & 63.94±1.34 & 55.92±1.26 & 6.8 \\
    % Last FA & 70.05±2.03 & 71.02±0.96 & 26.47±1.20 & 48.98±0.95 \
    % Every FA & 70.45±1.96 & 60.04±0.93 & 18.33±1.04 & 48.17±0.80 \
    & DIGL & \textit{79.70±2.15} & 70.76±0.77 & \underline{35.72±1.12} & \underline{69.76±0.42} & \underline{69.37±0.43} & 3.0 \\
    & SDRF & 71.05±1.87 & 70.92±0.79 & \textit{28.37±1.17} & 68.21±0.43 & 66.78±0.44 & 4.8 \\
    & FoSR & \underline{80.00±1.57} & \underline{73.42±0.81} & 25.07±0.99 & 57.27±0.54 & 56.82±0.60 & 4.6 \\
    & GTR & 79.10±1.86 & \textit{72.59±2.48} & 27.52±0.99 & \textit{69.37±0.38} & \textit{67.97±0.47} & 3.6 \\
    & GCN-GUMP & \bf{84.89±1.63} & \bf{74.88±0.87} & \bf{36.02±1.43} & \bf{77.97±0.42} & \bf{75.85±0.44} & 1.0 \\
    \midrule
    \multirow{7}{*}{GIN} & None & 77.70±3.60 & 70.80±0.83 & 33.80±1.12 & \textit{75.65±0.49} & 74.93±0.46 & 4.0 \\
    & None (+layer) & 69.80±2.75 & 68.71±0.96 & 25.92±1.07 & 73.49±0.46 & 72.47±0.53 & 6.6 \\
    % Last FA & & & & \
    % Every FA & & & & \
    & DIGL & \underline{79.80±2.08} & 70.71±0.67 & \textit{35.74±1.20} & \underline{79.37±0.43} & \underline{76.88±0.39} & 2.8 \\
    & SDRF & \textit{78.40±2.80} & 69.81±0.79 & \underline{35.82±1.09} & 74.55±0.54 & 73.89±0.43 & 4.2 \\
    & FoSR & 78.00±2.22 & \underline{75.11±0.82} & 29.20±1.38 & 70.15±0.47 & 69.93±0.45 & 5.2 \\
    & GTR & 77.60±2.84 & \textit{73.13±0.69} & 30.57±1.42 & 75.45±0.44 & \textit{75.28±0.42} & 4.2 \\
    & GIN-GUMP & \bf{86.72±1.53} & \bf{75.43±0.70} & \bf{48.43±1.24} & \bf{81.25±0.37} & \bf{78.45±0.44} & 1.0 \\
    \bottomrule
    \end{tabular}
    % \vspace{-0.5cm}
\end{table}

\begin{figure}[H]
   \centering
   \subfigure[]{
     \includegraphics[width=0.45\textwidth]{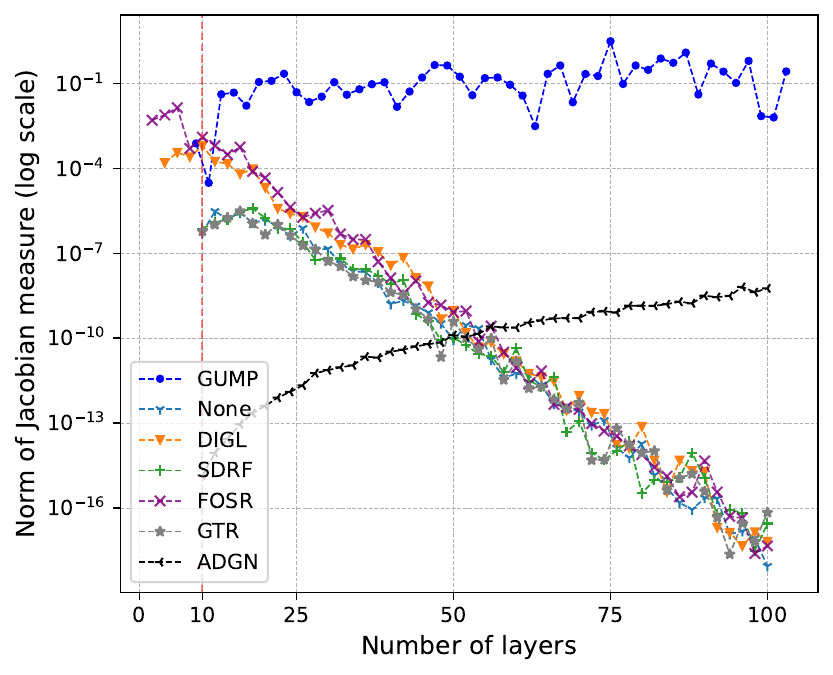}
     \label{fig:app-jac-vis}
  }
   \subfigure[]{
     \includegraphics[width=0.45\textwidth]{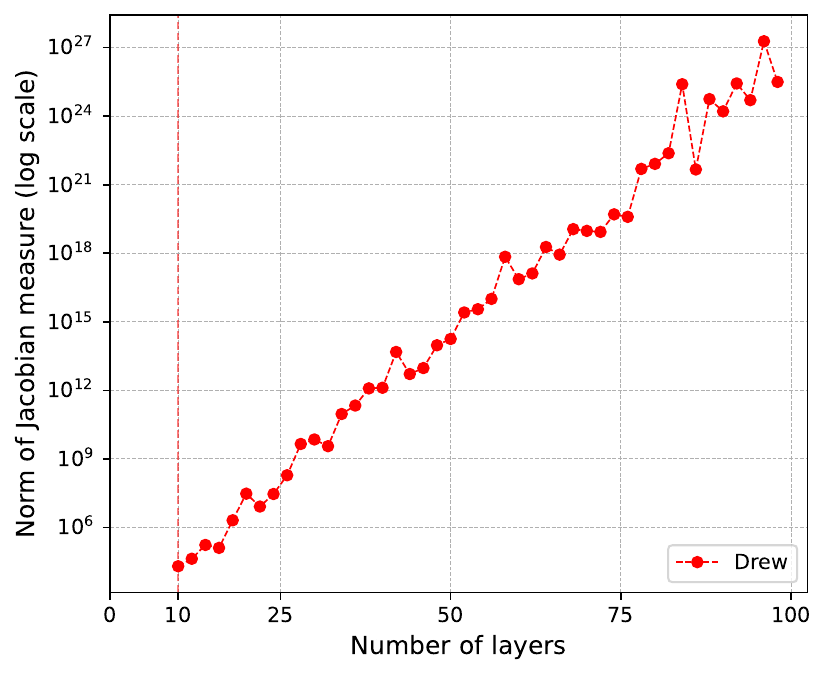}
     \label{fig:app-jac-vis-drew}
  }

   % \includegraphics[width=0.45\columnwidth]{img/jac_measure.pdf}
   % \missingfigure[figwidth=\columnwidth]{GUMP architecture}
   \vspace{-10px}
   \caption{Jacobian norm versus layers on NCI1} \label{fig:app-app_ablation}
\end{figure}

We also compare the Jacobian of GUMP with Drew and ADGN in Figure~\ref{fig:app-app_ablation}. Jacobian of Drew exponentially increases, suggesting its potential numerical instability when training Drew with more layers.
The Jacobian of ADGN is small when the ADGN layer is small and steadily increases to $10^{-8}$
as the ADGN layer reaches 100. Although the Jacobian of ADGN does not exhibit exponential decay, the correlation between distant nodes is significantly weaker compared to GUMP.

\subsection{OGB MOLHIV Results}\label{app:ogb-molhiv}

\paragraph{Setup}
We further evaluate graph classification on \texttt{ogbg-molhiv} from the Open Graph Benchmark (OGB)~\citep{hu2020open}. The dataset contains molecular graphs and uses scaffold splitting to test generalization to structurally different molecules. Following the official OGB protocol, we report test ROC-AUC.

\paragraph{Results}
Table~\ref{tab:ogb-molhiv} reports the OGB MOLHIV results with the same comparison methods as Table~\ref{tab:graph-classification-long-range}. GIN-GUMP achieves a test ROC-AUC of 77.89, outperforming GRIT at 77.07, GPS at 76.62, GTR at 76.07, and the reproduced Kiani et al. baseline at 75.48. These results provide an additional standardized graph-classification evaluation beyond the TU datasets and show that GUMP remains effective under the official scaffold split.

\begin{table}[H]
  \centering
  \caption{Experimental results on \texttt{ogbg-molhiv}. Test ROC-AUC (\%) is reported under the official scaffold split and evaluator.}
  \label{tab:ogb-molhiv}
  \setlength\tabcolsep{4pt}
  \begin{tabular}{c|l|c}
    \toprule
    Classes & Method & Test ROC-AUC (\%) $\uparrow$ \\ \midrule
    \multirow{1}{*}{--} & GCN & 74.47±1.51 \\ \midrule
    \multirow{4}{*}{\makecell{Rewiring}} & DIGL & 75.16±1.81 \\
    & SDRF & 75.00±1.08 \\
    & FoSR & 73.42±1.35 \\
    & GTR & 76.07±0.70 \\ \midrule
    \multirow{2}{*}{Diffusion} & ADGN & 73.21±0.21 \\
    & GRAND & 73.29±1.12 \\ \midrule
    Transformer & GPS & 76.62±0.82 \\ \midrule
    -- & Ortho-GConv & 70.96±0.64 \\ \midrule
    -- & Kiani et al. & 75.48±0.42 \\ \midrule
    K-hop MP & GRIT & 77.07±0.70 \\ \midrule
    Ours & GIN-GUMP & 77.89±0.54 \\ \bottomrule
  \end{tabular}
\end{table}

\subsection{Node Classification Results}\label{app:node-classification}
We also apply GUMP to node classification tasks on Cora and Citeseer datasets. The results are shown in Table~\ref{tab:node-classification}.

% \usepackage{multirow}
\begin{table}[H]
   \centering
  %  \scriptsize
   \vspace{-0.2cm}
   \caption{Accuracy of node classification datasets: Cora and Citeseer}\label{tab:node-classification}
   \begin{tabular}{c|c|ccccc}
      \toprule
   \multicolumn{2}{c|}{Layers}           & 2    & 4    & 8    & 16   & 64   \\\midrule
   \multirow{3}{*}{Cora}     & GCN      & 81.1 & 80.4 & 69.5 & 60.3 & 28.7 \\ 
                             & GCNII    & 82.2 & 82.6 & 84.2 & 84.6 & 85.5 \\
                             & GCN-GUMP & 84.6 & 86.2 & 84.8 & 85.4 & 87.4 \\ \midrule
   \multirow{3}{*}{Citeseer} & GCN      & 70.8 & 67.6 & 30.2 & 18.3 & 20.0 \\
                             & GCNII    & 68.2 & 68.9 & 70.6 & 72.9 & 73.4 \\
                             & GCN-GUMP & 73.0 & 73.0 & 72.8 & 72.4 & 75.8 \\ \bottomrule
   \end{tabular}
   \vspace{-0.2cm}
\end{table}

\paragraph{Training time of GUMP} The time of training GCN and GCN-GUMP for 100 epochs on the TUDataset is shown in Table~\ref{tab:time}. Using the same graph-level batch size as GCN is a practical setting for graph classification, but it is not a node-budget-normalized throughput comparison. Since GUMP propagates on the directed line graph, a graph-level mini-batch contains $\sum_g 2|E_g|$ line-graph nodes rather than $\sum_g |V_g|$ original graph nodes. Therefore, we also report a node-budget-matched GUMP setting with
\[
B_{\mathrm{GUMP}}^{\mathrm{match}}
=
\max\!\left(1,\left\lfloor B_{\mathrm{GCN}}\frac{\bar n}{2\bar m}\right\rfloor\right),
\]
where $\bar n$ and $\bar m$ are the average numbers of nodes and edges per graph. This setting controls the expected number of propagated nodes per mini-batch, making the comparison fairer in terms of the amount of graph state propagated in each update.

\begin{table}[H]
    \vspace{-0.2cm}
    \centering
    \caption{Training seconds on TUDataset for 100 epochs. The same-batch setting uses the shared GCN/GUMP graph-level batch sizes from the accuracy experiments, while the matched setting controls the expected propagated-node budget per mini-batch.}\label{tab:time}\label{tab:runtime}
    \footnotesize
    \setlength\tabcolsep{3pt}
    \begin{tabular}{c|ccccc}
        \toprule
        Dataset & \makecell{GCN\\time (s)} & \makecell{GCN/GUMP\\graph batch} & \makecell{Same-batch\\GUMP time (s)} & \makecell{Matched\\GUMP batch} & \makecell{Matched\\GUMP time (s)} \\ \midrule
        MUTAG    & 4.39  & 16 & 7.21  & 3 & 37.9  \\
        Proteins & 20.57 & 64 & 26.19 & 8 & 203.7 \\
        Enzymes  & 11.26 & 16 & 20.29 & 2 & 160.2 \\
        NCI1     & 71.79 & 16 & 82.02 & 7 & 187.7 \\
        NCI109   & 74.56 & 16 & 93.38 & 7 & 213.5 \\ \bottomrule
    \end{tabular}
    \vspace{-0.2cm}
\end{table}

Under this matched propagated-node budget, GUMP is more expensive than GCN: the matched GUMP times on MUTAG/PROTEINS/ENZYMES/NCI1/NCI109 are 37.9/203.7/160.2/187.7/213.5 seconds, respectively, which are larger than the corresponding GCN times. Thus, Table~\ref{tab:time} should be interpreted as showing the practical same-batch runtime and the additional compute cost incurred by GUMP under a fairer propagated-node budget, rather than as evidence of comparable per-node cost to GCN.

% \vspace{-0.3cm}
% \paragraph{Results following \citet{tonshoff2023did}} We compare GCN-GUMP with other GNNs on Peptides-func and Peptides-struct datasets following \citet{tonshoff2023did}. The results are shown in Table~\ref{table:peptides}. The results show that GCN-GUMP ranks second on Peptides-func and ranks first on Peptides-struct as a one-hop message passing methods.

% \paragraph{Average time of preprocessing} The average time of preprocessing for line graph on various datasets is shown in Table~\ref{tab:preprocess-time}.
    
% \begin{table}[h]
%     \centering
%     % \scriptsize
%     \setlength\tabcolsep{4pt}
%     \begin{tabular}{c|ccccccc}
%     \toprule
%                  & MUTAG & Proteins & Enzymes & NCI1 & NCI109 & Peptides-func & Peptides-struct \\
%     \midrule
%     Avg. Time    & 0.001 & 0.005    & 0.009   & 0.002 & 0.002  & 0.014         & 0.013           \\
%     \bottomrule
%     \end{tabular}
%     \caption{Average times (seconds) of preprocessing line graph for various datasets}
%     \label{tab:preprocess-time}
% \end{table}

\clearpage

\section{Apply GUMP to Directed Graphs}\label{app:direct_graph}
GUMP can also be applied to directed graphs as shown in Algorithm~\ref{alg:dig-transformation}. The transformation of directed graphs is also based on Lemma~\ref{lem:eulerian-graph}.

\begin{algorithm}[H]
  % \small
	\caption{Graph transformation for directed graphs}
	\begin{algorithmic}[1]
		\REQUIRE A directed graph $G=(V,E)$;
        \STATE Initialize a new digraph $G^\prime = (V, E^\prime)$;
		\FOR{$(i,j)\in E$}
        \STATE Add $(i,j)$ and $(j,i)$ to $E^\prime$;
		\ENDFOR
        \STATE Remove duplicated edges in $E^\prime$;
        % \STATE remove duplicated edges in $E^\prime$;
        \STATE Convert $G^\prime$ to its line graph $\mathsf{L}(G^\prime)$;
		\STATE \textbf{Return:} A digraph $\mathsf{L}(G^\prime)$.
	\end{algorithmic}
	\label{alg:dig-transformation}
\end{algorithm}

\end{document}

%% file: math_commands.tex
\usepackage{amsmath,amsfonts,bm}

\def\eqref#1{equation~\ref{#1}}
\def\1{\bm{1}}

\DeclareMathAlphabet{\mathsfit}{\encodingdefault}{\sfdefault}{m}{sl}
\SetMathAlphabet{\mathsfit}{bold}{\encodingdefault}{\sfdefault}{bx}{n}

%% file: main.bib
@InProceedings{li2018deeper,
  author    = {Li, Qimai and Han, Zhichao and Wu, Xiao-Ming},
  booktitle = {AAAI Conference on Artificial Intelligence},
  title     = {Deeper insights into graph convolutional networks for semi-supervised learning},
  year      = {2018},
  pages     = {3538--3545},
  volume    = {32},
}

@InProceedings{li2019deepgcns,
  author    = {Li, Qimai and Han, Zhichao and Wu, Xiao-Ming},
  booktitle = {IEEE Conference on Computer Vision and Pattern Recognition},
  title     = {Deepgcns: Can gcns go as deep as cnns?},
  year      = {2019},
  pages     = {9267--9276},
}

@InProceedings{chen2020measuring,
  author    = {Chen, Jianwen and He, Jun and Chen, Changyou and Wang, Shuiwang},
  booktitle = {AAAI Conference on Artificial Intelligence},
  title     = {Measuring and Relieving the Over-smoothing Problem for Graph Neural Networks from the Topological View},
  year      = {2020},
  number    = {07},
  pages     = {12492--12499},
  volume    = {34},
}

@article{alon2020bottleneck,
  title={On the bottleneck of graph neural networks and its practical implications},
  author={Alon, Uri and Yahav, Eran},
  journal={arXiv preprint arXiv:2006.05205},
  year={2020}
}

@inproceedings{topping2021understanding,
  title={Understanding over-squashing and bottlenecks on graphs via curvature},
  author={Topping, Jake and Di Giovanni, Francesco and Chamberlain, Benjamin Paul and Dong, Xiaowen and Bronstein, Michael M},
  booktitle={International Conference on Learning Representations},
  year={2022}
}

@article{severini2003digraph,
  title={On the digraph of a unitary matrix},
  author={Severini, Simone},
  journal={SIAM Journal on Matrix Analysis and Applications},
  volume={25},
  number={1},
  pages={295--300},
  year={2003},
  publisher={SIAM}
}

@article{kiani2022projunn,
  title={projUNN: Efficient method for training deep networks with unitary matrices},
  author={Kiani, Bobak and Balestriero, Randall and LeCun, Yann and Lloyd, Seth},
  journal={Advances in Neural Information Processing Systems},
  volume={35},
  pages={14448--14463},
  year={2022}
}

@inproceedings{kiani2024unitary,
  title={Unitary convolutions for learning on graphs and groups},
  author={Kiani, Bobak T. and Fesser, Lukas and Weber, Melanie},
  booktitle={Advances in Neural Information Processing Systems},
  volume={37},
  pages={136922--136961},
  doi={10.52202/079017-4351},
  eprint={2410.05499},
  archivePrefix={arXiv},
  year={2024}
}

@inproceedings{black2023understanding,
  title={Understanding oversquashing in gnns through the lens of effective resistance},
  author={Black, Mitchell and Wan, Zhengchao and Nayyeri, Amir and Wang, Yusu},
  booktitle={International Conference on Machine Learning},
  pages={2528--2547},
  year={2023},
  organization={PMLR}
}

@inproceedings{gilmer2017neural,
  title={Neural message passing for quantum chemistry},
  author={Gilmer, Justin and Schoenholz, Samuel S and Riley, Patrick F and Vinyals, Oriol and Dahl, George E},
  booktitle={International conference on machine learning},
  pages={1263--1272},
  year={2017},
  organization={PMLR}
}

@inproceedings{arnaiz2022diffwire,
title={DiffWire: Inductive Graph Rewiring via the Lov\'asz Bound},
author={Adri{\'a}n Arnaiz-Rodr{\'\i}guez and Ahmed Begga and Francisco Escolano and Nuria M Oliver},
booktitle={The First Learning on Graphs Conference},
year={2022},
url={https://openreview.net/forum?id=IXvfIex0mX6f}
}

@inproceedings{xu2018representation,
  title={Representation learning on graphs with jumping knowledge networks},
  author={Xu, Keyulu and Li, Chengtao and Tian, Yonglong and Sonobe, Tomohiro and Kawarabayashi, Ken-ichi and Jegelka, Stefanie},
  booktitle={International conference on machine learning},
  pages={5453--5462},
  year={2018},
  organization={PMLR}
}

@inproceedings{jing2017tunable,
  title={Tunable efficient unitary neural networks (eunn) and their application to rnns},
  author={Jing, Li and Shen, Yichen and Dubcek, Tena and Peurifoy, John and Skirlo, Scott and LeCun, Yann and Tegmark, Max and Solja{\v{c}}i{\'c}, Marin},
  booktitle={International Conference on Machine Learning},
  pages={1733--1741},
  year={2017},
  organization={PMLR}
}

@inproceedings{arjovsky2016unitary,
  title={Unitary evolution recurrent neural networks},
  author={Arjovsky, Martin and Shah, Amar and Bengio, Yoshua},
  booktitle={International conference on machine learning},
  pages={1120--1128},
  year={2016},
  organization={PMLR}
}

@article{keller1975closest,
  title={Closest unitary, orthogonal and hermitian operators to a given operator},
  author={Keller, Joseph B},
  journal={Mathematics Magazine},
  volume={48},
  number={4},
  pages={192--197},
  year={1975},
  publisher={JSTOR}
}

@InProceedings{banerjee2022oversquashing,
  author       = {Banerjee, Pradeep Kr and Karhadkar, Kedar and Wang, Yu Guang and Alon, Uri and Mont{\'u}far, Guido},
  booktitle    = {Annual Allerton Conference on Communication, Control, and Computing (Allerton)},
  title        = {Oversquashing in GNNs through the lens of information contraction and graph expansion},
  year         = {2022},
  organization = {IEEE},
  pages        = {1--8},
}

@inproceedings{di2023over,
  title={On over-squashing in message passing neural networks: The impact of width, depth, and topology},
  author={Di Giovanni, Francesco and Giusti, Lorenzo and Barbero, Federico and Luise, Giulia and Lio, Pietro and Bronstein, Michael M},
  booktitle={International Conference on Machine Learning},
  pages={7865--7885},
  year={2023},
  organization={PMLR}
}

@article{barbero2023locality,
  title={Locality-Aware Graph-Rewiring in GNNs},
  author={Barbero, Federico and Velingker, Ameya and Saberi, Amin and Bronstein, Michael and Di Giovanni, Francesco},
  journal={arXiv preprint arXiv:2310.01668},
  year={2023}
}

@book{bang2008digraphs,
  title={Digraphs: theory, algorithms and applications},
  author={Bang-Jensen, J{\o}rgen and Gutin, Gregory Z},
  year={2008},
  publisher={Springer Science \& Business Media}
}

@inproceedings{helfrich2018orthogonal,
  title={Orthogonal recurrent neural networks with scaled Cayley transform},
  author={Helfrich, Kyle and Willmott, Devin and Ye, Qiang},
  booktitle={International Conference on Machine Learning},
  pages={1969--1978},
  year={2018},
  organization={PMLR}
}

@inproceedings{lezcano2019cheap,
  title={Cheap orthogonal constraints in neural networks: A simple parametrization of the orthogonal and unitary group},
  author={Lezcano-Casado, Mario and Mart{\i}nez-Rubio, David},
  booktitle={International Conference on Machine Learning},
  pages={3794--3803},
  year={2019},
  organization={PMLR}
}

@article{sedghi2018singular,
  title={The singular values of convolutional layers},
  author={Sedghi, Hanie and Gupta, Vineet and Long, Philip M},
  journal={arXiv preprint arXiv:1805.10408},
  year={2018}
}

@article{li2019preventing,
  title={Preventing gradient attenuation in lipschitz constrained convolutional networks},
  author={Li, Qiyang and Haque, Saminul and Anil, Cem and Lucas, James and Grosse, Roger B and Jacobsen, J{\"o}rn-Henrik},
  journal={Advances in neural information processing systems},
  volume={32},
  year={2019}
}

@inproceedings{singla2021skew,
  title={Skew orthogonal convolutions},
  author={Singla, Sahil and Feizi, Soheil},
  booktitle={International Conference on Machine Learning},
  pages={9756--9766},
  year={2021},
  organization={PMLR}
}

@article{trockman2021orthogonalizing,
  title={Orthogonalizing convolutional layers with the cayley transform},
  author={Trockman, Asher and Kolter, J Zico},
  journal={arXiv preprint arXiv:2104.07167},
  year={2021}
}

@InProceedings{guo2022orthogonal,
  author    = {Guo, Kai and Zhou, Kaixiong and Hu, Xia and Li, Yu and Chang, Yi and Wang, Xin},
  booktitle = {AAAI Conference on Artificial Intelligence},
  title     = {Orthogonal graph neural networks},
  year      = {2022},
  number    = {4},
  pages     = {3996--4004},
  volume    = {36},
}

@inproceedings{pascanu2013difficulty,
  title={On the difficulty of training recurrent neural networks},
  author={Pascanu, Razvan and Mikolov, Tomas and Bengio, Yoshua},
  booktitle={International conference on machine learning},
  pages={1310--1318},
  year={2013},
  organization={Pmlr}
}

@inproceedings{morris2020tudataset,
    title={TUDataset: A collection of benchmark datasets for learning with graphs},
    author={Christopher Morris and Nils M. Kriege and Franka Bause and Kristian Kersting and Petra Mutzel and Marion Neumann},
    booktitle={ICML 2020 Workshop on Graph Representation Learning and Beyond (GRL+ 2020)},
    archivePrefix={arXiv},
    eprint={2007.08663},
    url={www.graphlearning.io},
    year={2020}
}

@article{dwivedi2022long,
  title={Long range graph benchmark},
  author={Dwivedi, Vijay Prakash and Ramp{\'a}{\v{s}}ek, Ladislav and Galkin, Michael and Parviz, Ali and Wolf, Guy and Luu, Anh Tuan and Beaini, Dominique},
  journal={Advances in Neural Information Processing Systems},
  volume={35},
  pages={22326--22340},
  year={2022}
}

@inproceedings{hu2020open,
  title={Open Graph Benchmark: Datasets for Machine Learning on Graphs},
  author={Hu, Weihua and Fey, Matthias and Zitnik, Marinka and Dong, Yuxiao and Ren, Hongyu and Liu, Bowen and Catasta, Michele and Leskovec, Jure},
  booktitle={Advances in Neural Information Processing Systems},
  volume={33},
  pages={22118--22133},
  year={2020}
}

@InProceedings{morris2019weisfeiler,
  author    = {Morris, Christopher and Ritzert, Martin and Fey, Matthias and Hamilton, William L and Lenssen, Jan Eric and Rattan, Gaurav and Grohe, Martin},
  booktitle = {AAAI conference on artificial intelligence},
  title     = {Weisfeiler and leman go neural: Higher-order graph neural networks},
  year      = {2019},
  number    = {01},
  pages     = {4602--4609},
  volume    = {33},
}

@article{battaglia2018relational,
  title={Relational inductive biases, deep learning, and graph networks},
  author={Battaglia, Peter W and Hamrick, Jessica B and Bapst, Victor and Sanchez-Gonzalez, Alvaro and Zambaldi, Vinicius and Malinowski, Mateusz and Tacchetti, Andrea and Raposo, David and Santoro, Adam and Faulkner, Ryan and others},
  journal={arXiv preprint arXiv:1806.01261},
  year={2018}
}

@article{gasteiger2019diffusion,
  title={Diffusion improves graph learning},
  author={Gasteiger, Johannes and Wei{\ss}enberger, Stefan and G{\"u}nnemann, Stephan},
  journal={Advances in neural information processing systems},
  volume={32},
  year={2019}
}

@inproceedings{fey2019pyg,
  title={Fast Graph Representation Learning with {PyTorch Geometric}},
  author={Fey, Matthias and Lenssen, Jan E.},
  booktitle={ICLR Workshop on Representation Learning on Graphs and Manifolds},
  year={2019},
}

@inproceedings{orvieto2023resurrecting,
  title={Resurrecting recurrent neural networks for long sequences},
  author={Orvieto, Antonio and Smith, Samuel L and Gu, Albert and Fernando, Anushan and Gulcehre, Caglar and Pascanu, Razvan and De, Soham},
  booktitle={International Conference on Machine Learning},
  pages={26670--26698},
  year={2023},
  organization={PMLR}
}

@inproceedings{chamberlain2021grand,
  title={Grand: Graph neural diffusion},
  author={Chamberlain, Ben and Rowbottom, James and Gorinova, Maria I and Bronstein, Michael and Webb, Stefan and Rossi, Emanuele},
  booktitle={International Conference on Machine Learning},
  pages={1407--1418},
  year={2021},
  organization={PMLR}
}

@article{de2024griffin,
  title={Griffin: Mixing Gated Linear Recurrences with Local Attention for Efficient Language Models},
  author={De, Soham and Smith, Samuel L and Fernando, Anushan and Botev, Aleksandar and Cristian-Muraru, George and Gu, Albert and Haroun, Ruba and Berrada, Leonard and Chen, Yutian and Srinivasan, Srivatsan and others},
  journal={arXiv preprint arXiv:2402.19427},
  year={2024}
}

@inproceedings{gutteridge2023drew,
  title={Drew: Dynamically rewired message passing with delay},
  author={Gutteridge, Benjamin and Dong, Xiaowen and Bronstein, Michael M and Di Giovanni, Francesco},
  booktitle={International Conference on Machine Learning},
  pages={12252--12267},
  year={2023},
  organization={PMLR}
}

@article{peng2023rwkv,
  title={Rwkv: Reinventing rnns for the transformer era},
  author={Peng, Bo and Alcaide, Eric and Anthony, Quentin and Albalak, Alon and Arcadinho, Samuel and Biderman, Stella and Cao, Huanqi and Cheng, Xin and Chung, Michael and Grella, Matteo and others},
  journal={arXiv preprint arXiv:2305.13048},
  year={2023}
}

@article{gu2023mamba,
  title={Mamba: Linear-time sequence modeling with selective state spaces},
  author={Gu, Albert and Dao, Tri},
  journal={arXiv preprint arXiv:2312.00752},
  year={2023}
}

@article{tonshoff2023did,
  title={Where did the gap go? reassessing the long-range graph benchmark},
  author={T{\"o}nshoff, Jan and Ritzert, Martin and Rosenbluth, Eran and Grohe, Martin},
  journal={arXiv preprint arXiv:2309.00367},
  year={2023}
}

@Misc{jordan2024muon,
  author = {Keller Jordan and Yuchen Jin and Vlado Boza and You Jiacheng and Franz Cesista and Laker Newhouse and Jeremy Bernstein},
  title  = {Muon: An optimizer for hidden layers in neural networks},
  year   = {2024},
}

@article{saxe2013exact,
  title={Exact solutions to the nonlinear dynamics of learning in deep linear neural networks},
  author={Saxe, Andrew M and McClelland, James L and Ganguli, Surya},
  journal={arXiv preprint arXiv:1312.6120},
  year={2013}
}

@article{liu2025muon,
  title={Muon is scalable for LLM training},
  author={Liu, Jingyuan and Su, Jianlin and Yao, Xingcheng and Jiang, Zhejun and Lai, Guokun and Du, Yulun and Qin, Yidao and Xu, Weixin and Lu, Enzhe and Yan, Junjie and others},
  journal={arXiv preprint arXiv:2502.16982},
  year={2025}
}

@article{rampavsek2022recipe,
  title={Recipe for a general, powerful, scalable graph transformer},
  author={Ramp{\'a}{\v{s}}ek, Ladislav and Galkin, Michael and Dwivedi, Vijay Prakash and Luu, Anh Tuan and Wolf, Guy and Beaini, Dominique},
  journal={Advances in Neural Information Processing Systems},
  volume={35},
  pages={14501--14515},
  year={2022}
}

@inproceedings{shirzad2023exphormer,
  title={Exphormer: Sparse transformers for graphs},
  author={Shirzad, Hamed and Velingker, Ameya and Venkatachalam, Balaji and Sutherland, Danica J and Sinop, Ali Kemal},
  booktitle={International Conference on Machine Learning},
  pages={31613--31632},
  year={2023},
  organization={PMLR}
}

@inproceedings{he2023generalization,
  title={A generalization of vit/mlp-mixer to graphs},
  author={He, Xiaoxin and Hooi, Bryan and Laurent, Thomas and Perold, Adam and LeCun, Yann and Bresson, Xavier},
  booktitle={International conference on machine learning},
  pages={12724--12745},
  year={2023},
  organization={PMLR}
}

@article{oono2019graph,
  title={Graph neural networks exponentially lose expressive power for node classification},
  author={Oono, Kenta and Suzuki, Taiji},
  journal={arXiv preprint arXiv:1905.10947},
  year={2019}
}

@article{wu2020comprehensive,
  title={A comprehensive survey on graph neural networks},
  author={Wu, Zonghan and Pan, Shirui and Chen, Fengwen and Long, Guodong and Zhang, Chengqi and Yu, Philip S},
  journal={IEEE transactions on neural networks and learning systems},
  volume={32},
  number={1},
  pages={4--24},
  year={2020},
  publisher={IEEE}
}

@article{zhao2020pairnorm,
  title={Pairnorm: Tackling oversmoothing in gnns},
  author={Zhao, Lingxiao and Akoglu, Leman},
  journal={arXiv preprint arXiv:1909.12223},
  year={2019}
}

@inproceedings{chen2020simple,
  title={Simple and deep graph convolutional networks},
  author={Chen, Ming and Wei, Zhewei and Huang, Zengfeng and Ding, Bolin and Li, Yaliang},
  booktitle={International conference on machine learning},
  pages={1725--1735},
  year={2020},
  organization={PMLR}
}

@inproceedings{gao2019graph,
  title={Graph u-nets},
  author={Gao, Hongyang and Ji, Shuiwang},
  booktitle={international conference on machine learning},
  pages={2083--2092},
  year={2019},
  organization={PMLR}
}

@inproceedings{balduzzi2017shattered,
  title={The shattered gradients problem: If resnets are the answer, then what is the question?},
  author={Balduzzi, David and Frean, Marcus and Leary, Lennox and Lewis, JP and Ma, Kurt Wan-Duo and McWilliams, Brian},
  booktitle={International conference on machine learning},
  pages={342--350},
  year={2017},
  organization={PMLR}
}

@article{arroyo2025vanishing,
  title={On vanishing gradients, over-smoothing, and over-squashing in gnns: Bridging recurrent and graph learning},
  author={Arroyo, {\'A}lvaro and Gravina, Alessio and Gutteridge, Benjamin and Barbero, Federico and Gallicchio, Claudio and Dong, Xiaowen and Bronstein, Michael and Vandergheynst, Pierre},
  journal={arXiv preprint arXiv:2502.10818},
  year={2025}
}

@article{jamadandi2024spectral,
  title={Spectral graph pruning against over-squashing and over-smoothing},
  author={Jamadandi, Adarsh and Rubio-Madrigal, Celia and Burkholz, Rebekka},
  journal={Advances in Neural Information Processing Systems},
  volume={37},
  pages={10348--10379},
  year={2024}
}

@article{kovarik1970some,
  title={Some iterative methods for improving orthonormality},
  author={Kovarik, Zdislav},
  journal={SIAM Journal on Numerical Analysis},
  volume={7},
  number={3},
  pages={386--389},
  year={1970},
  publisher={SIAM}
}

@article{bjorck1971iterative,
  title={An iterative algorithm for computing the best estimate of an orthogonal matrix},
  author={Bj{\"o}rck, {\AA}ke and Bowie, Clazett},
  journal={SIAM Journal on Numerical Analysis},
  volume={8},
  number={2},
  pages={358--364},
  year={1971},
  publisher={SIAM}
}

@Article{kipf2016semi,
  author  = {Kipf, Thomas N and Welling, Max},
  journal = {arXiv preprint arXiv:1609.02907},
  title   = {Semi-supervised classification with graph convolutional networks},
  year    = {2016},
}

@Article{rong2019dropedge,
  author  = {Rong, Yu and Huang, Wenbing and Xu, Tingyang and Huang, Junzhou},
  journal = {arXiv preprint arXiv:1907.10903},
  title   = {Dropedge: Towards deep graph convolutional networks on node classification},
  year    = {2019},
}

@InProceedings{Karhadkar2023,
  author    = {Karhadkar, Kedar and Banerjee, Pradeep and Montufar, Guido},
  booktitle = {International Conference on Learning Representations},
  title     = {FoSR: First-order spectral rewiring for addressing oversquashing in GNNs},
  year      = {2023},
}

@article{cai2022line,
  title   = {Line Graph Neural Networks for Link Prediction},
  author  = {Cai, Lei and Li, Jundong and Wang, Jie and Ji, Shuiwang},
  journal = {IEEE Transactions on Pattern Analysis and Machine Intelligence},
  volume  = {44},
  number  = {9},
  pages   = {5103--5113},
  year    = {2022},
  doi     = {10.1109/TPAMI.2021.3080635}
}

@article{zhang2023line,
  title   = {Line Graph Contrastive Learning for Link Prediction},
  author  = {Zhang, Zehua and Sun, Shilin and Ma, Guixiang and Zhong, Caiming},
  journal = {Pattern Recognition},
  volume  = {140},
  pages   = {109537},
  year    = {2023},
  doi     = {10.1016/j.patcog.2023.109537}
}

@inproceedings{jiang2019censnet,
  title     = {CensNet: Convolution with Edge-Node Switching in Graph Neural Networks},
  author    = {Jiang, Xiaodong and Ji, Pengsheng and Li, Sheng},
  booktitle = {Proceedings of the Twenty-Eighth International Joint Conference on Artificial Intelligence},
  pages     = {2656--2662},
  year      = {2019},
  doi       = {10.24963/ijcai.2019/369}
}

@article{monti2018dual,
  title         = {Dual-Primal Graph Convolutional Networks},
  author        = {Monti, Federico and Shchur, Oleksandr and Bojchevski, Aleksandar and Litany, Or and G{\"u}nnemann, Stephan and Bronstein, Michael M.},
  journal       = {arXiv preprint arXiv:1806.00770},
  year          = {2018},
  eprint        = {1806.00770},
  archivePrefix = {arXiv}
}

@inproceedings{yang2020nenn,
  title     = {NENN: Incorporate Node and Edge Features in Graph Neural Networks},
  author    = {Yang, Yulei and Li, Dongsheng},
  booktitle = {Proceedings of The 12th Asian Conference on Machine Learning},
  series    = {Proceedings of Machine Learning Research},
  volume    = {129},
  pages     = {593--608},
  year      = {2020},
  publisher = {PMLR}
}

@inproceedings{wang2021egat,
  title     = {EGAT: Edge-Featured Graph Attention Network},
  author    = {Wang, Ziming and Chen, Jun and Chen, Haopeng},
  booktitle = {Artificial Neural Networks and Machine Learning -- ICANN 2021},
  series    = {Lecture Notes in Computer Science},
  pages     = {253--264},
  year      = {2021},
  publisher = {Springer},
  doi       = {10.1007/978-3-030-86362-3_21}
}

@inproceedings{zhang2018link,
  title     = {Link Prediction Based on Graph Neural Networks},
  author    = {Zhang, Muhan and Chen, Yixin},
  booktitle = {Advances in Neural Information Processing Systems},
  volume    = {31},
  year      = {2018}
}

@inproceedings{pan2022neural,
  title     = {Neural Link Prediction with Walk Pooling},
  author    = {Pan, Liming and Shi, Cheng and Dokmani{\'c}, Ivan},
  booktitle = {International Conference on Learning Representations},
  year      = {2022}
}

@inproceedings{chamberlain2023graph,
  title     = {Graph Neural Networks for Link Prediction with Subgraph Sketching},
  author    = {Chamberlain, Benjamin Paul and Shirobokov, Sergey and Rossi, Emanuele and Frasca, Fabrizio and Markovich, Thomas and Hammerla, Nils Yannick and Bronstein, Michael M. and Hansmire, Max},
  booktitle = {International Conference on Learning Representations},
  year      = {2023}
}

@article{zhang2025directed,
  title         = {Directed Link Prediction using GNN with Local and Global Feature Fusion},
  author        = {Zhang, Yuyang and Shen, Xu and Xie, Yu and Wong, Ka-Chun and Xie, Weidun and Peng, Chengbin},
  journal       = {arXiv preprint arXiv:2506.20235},
  year          = {2025},
  eprint        = {2506.20235},
  archivePrefix = {arXiv}
}
